\documentclass[lettersize,journal]{IEEEtran}
\usepackage{amsmath,amsfonts}
\usepackage{algorithmic}
\usepackage{algorithm}
\usepackage{array}
\usepackage[caption=false,font=normalsize,labelfont=sf,textfont=sf]{subfig}
\usepackage{textcomp}
\usepackage{stfloats}
\usepackage{url}
\usepackage{verbatim}
\usepackage{graphicx}
\usepackage{cite}
\usepackage{multirow}
\usepackage{booktabs}
\usepackage{xcolor}
\usepackage{colortbl}
\usepackage{orcidlink}
\usepackage{bbding}
\hypersetup{%
    pdfborder = {0 0 0}
}
\definecolor{mygray}{gray}{.9}
\definecolor{myblue}{RGB}{93,80,180}
\definecolor{mygreen}{RGB}{93,173,85}
\begin{document}

\title{LOGCAN++: ADAPTIVE LOCAL-GLOBAL CLASS-AWARE NETWORK FOR SEMANTIC
SEGMENTATION OF REMOTE SENSING IMAGES}
\author{Xiaowen Ma~\orcidlink{0000-0001-5031-2641}, 
Rongrong Lian \orcidlink{0009-0005-3262-6651}, 
Zhenkai Wu \orcidlink{0009-0000-0613-0584}, 
Hongbo Guo \orcidlink{0009-0003-6585-1471},
Fan Yang \orcidlink{0009-0002-1292-0894},
Mengting Ma~\orcidlink{0000-0002-6897-3576},
Sensen Wu~\orcidlink{0000-0001-9322-0149}, 
Zhenhong Du \orcidlink{0000-0001-9449-0415}, 
Wei Zhang~\orcidlink{0000-0002-4424-079X},
and  Siyang Song~\orcidlink{0000-0003-2339-5685} \vspace{.5em} \\
\small\textcolor{magenta}{\textbf{\url{https://github.com/xwmaxwma/rssegmentation}}}
\thanks{This work was supported by the Public Welfare Science and Technology Plan of Ningbo City (2022S125) and the Key Research and Development Plan of Zhejiang Province (2021C01031). \emph{(Xiaowen Ma and Rongrong Lian contribute equally. Corresponding author: Siyang Song and Wei Zhang.)}}

\thanks{Xiaowen Ma, Rongrong Lian, Zhenkai Wu 
 and Fan Yang are with the School of Software Technology, Zhejiang University, Hangzhou 310027, China (e-mail: xwma@zju.edu.cn).}

\thanks{Hongbo Guo is with the school of Electronic Information and Electrical Engineering, Shanghai Jiao Tong University, Shanghai 200030, China.}

\thanks{Mengting Ma is with the School of Computer Science and Technology, Zhejiang University, Hangzhou 310027, China.}

\thanks{Sensen Wu and Zhenhong Du are with the School of Earth Sciences, Zhejiang University, Hangzhou 310027, China.}
\thanks{Wei Zhang is with the School of Software Technology, Zhejiang University, Hangzhou 310027, China, and also with the Innovation Center of Yangtze River Delta, Zhejiang University, Jiaxing Zhejiang, 314103, China (e-mail: cstzhangwei@zju.edu.cn).}
\thanks{Siyang Song is with the HUBG Lab, Department of Computer Science, University of Exeter, UK (e-mail: s.song@exeter.ac.uk).}


}

\markboth{Submit to IEEE TRANSACTIONS ON GEOSCIENCE AND REMOTE SENSING}%
{Shell \MakeLowercase{\textit{et al.}}: A Sample Article Using IEEEtran.cls for IEEE Journals}


\maketitle

\begin{abstract}
Remote sensing images usually characterized by complex backgrounds, scale and orientation variations, and large intra-class variance. General semantic segmentation methods usually fail to fully investigate the above issues, and thus their performances on remote sensing image segmentation are limited. In this paper, we propose our LOGCAN++, a semantic segmentation model customized for remote sensing images, which is made up of a Global Class Awareness (GCA) module and several Local Class Awareness (LCA) modules. The GCA module captures global representations for class-level context modeling to reduce the interference of background noise. The LCA module generates local class representations as intermediate perceptual elements to indirectly associate pixels with the global class representations, targeting at dealing with the large intra-class variance problem. In particular, we introduce affine transformations in the LCA module for adaptive extraction of local class representations to effectively tolerate scale and orientation variations in remote sensing images. Extensive experiments on three benchmark datasets show that our LOGCAN++ outperforms current mainstream general and remote sensing semantic segmentation methods and achieves a better trade-off between speed and accuracy. 

\end{abstract}


\begin{figure*}[t]
	\centering \includegraphics[width=1.0\textwidth]
       {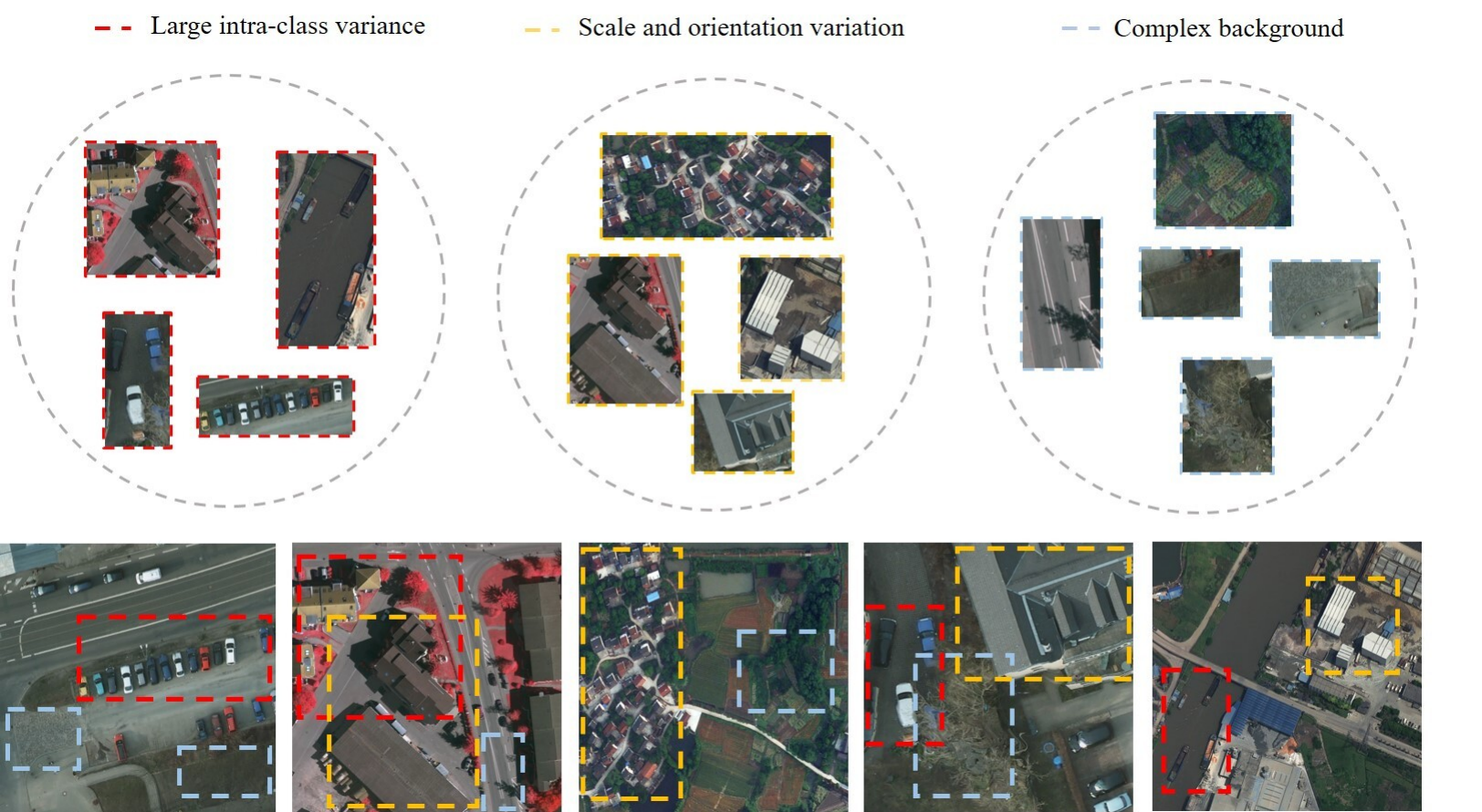}%
	\centering
	\caption{Visual examples of remote sensing image characteristics. The images at the bottom are selected from the ISPRS Vaihingen, ISPRS Potsdam and the LoveDA datasets. The images in the red, yellow and blue boxes represent the characteristics of remote sensing images with complex backgrounds, scale and orientation variations, and large intra-class variance, respectively.}
\label{fig:1}
\end{figure*}

\section{Introduction}

\IEEEPARstart{S}{emantic} segmentation of remote sensing images, aims to assign explicit categories to each pixel of a geospatial object in an image. This technique has a wide range of applications in fields such as environmental protection \cite{environment}, resource investigation~\cite{water}, and urban planning \cite{urban,urban2}. In recent years, the rapid development of remote sensing platforms and sensors has resulted in higher spatial, spectral and temporal resolution of remote sensing images \cite{rsdataset}. This provides the possibility of detailed observation of the Earth's surface features. As a result, more advanced semantic segmentation techniques for remote sensing images are constantly developed to meet the needs to process rich data \cite{9756442,10542207}.

Compared to natural images, semantic segmentation of remote sensing images is more challenging due to the following reasons: 

\textbf{\emph{1) Complex backgrounds.}} semantic segmentation of remote sensing images aims to segmenting geospatial objects such as buildings, roads, and rivers. These objects usually have diverse shapes, distributions and locations in different scenes, as illustrated in Fig. \ref{fig:1}. Extremely complex scenes can cause a large number of false alarms. 

\textbf{\emph{2) Large intra-class variance.}} For objects belong to the same category (e.g., buildings, roads and farmland), there is a large variance in the extracted semantic features due to differences in their sizes, shapes, colors, lighting conditions and even imaging conditions. In addition, complex backgrounds further increase their intra-class variance. In other words, large intra-class variance is detrimental to the recognition of semantic categories.

\textbf{\emph{3) Scale and orientation variations.}} Objects in natural images are generally upward due to gravity with small scale gaps. However, due to the overhead view and large imaging range, objects in remote sensing images usually show different orientations and suffer from large scale differences. Therefore, special contextual modeling techniques are needed to cope with such scale and orientation variations for remote sensing image segmentation.

In this paper, we focus on designing a effective network structure to address the aforementioned issues. We first analyze the convolution-based neural networks. Existing convolutional neural network (CNN)-based semantic segmentation methods focus on context modeling \cite{pspnet,nonlocal,danet,ccnet,caa}, which can be divided into spatial context modeling and relational context modeling. Spatial context modeling methods, such as PSPNet \cite{pspnet} and DeepLabv3+ \cite{deeplabv3+}, use spatial pyramid pooling (SPP) or empty spatial pyramid pooling (ASPP) to integrate spatial context information. Although these methods can capture homogeneous contextual dependencies, they ignore the differences between classes, which may produce unreliable contextual information when dealing with remote sensing images with complex features and large spectral differences. 
Relational context modeling methods capture heterogeneous dependencies based on attention mechanism. nonlocal neural network \cite{nonlocal} is proposed to compute the similarity of pixel pairs within an image for weighted aggregation, while DANet \cite{danet} uses spatial attention and channel attention for selective aggregation. However, such intensive attention operations are not applicable to remote sensing images, as their complex backgrounds would cause these operations to generate a large amount of background noises, leading to degraded semantic segmentation performances. Although recently proposed class-level context modeling approaches (e.g., ACFNet \cite{acfnet}, OCRNet \cite{ocrnet}, CCENet \cite{ccenet} and HMANet \cite{hmanet}) can reduce background noise interference caused by dense attention by integrating class-level contexts (i.e., class centers) using global class representations, they are still not perform well in remote sensing images. This is caused by the fact that remote sensing images have large intra-class variance, which leads to a large semantic gap between features and global class centers, and thus hindering models' performances.

Consequently, we turn our attention to transformer-based methods \cite{segmenter,segformer,unetformer,mask2former,biformer}. Since their core component is spatial self-attention, it also faces the problem of massive background noise interference when processing remote sensing images, and consequently leading to poor performance on most remote sensing datasets. In particular, due to the quadratic computational complexity of the spatial self-attention operation, it imposes heavy computational cost and memory consumption when applied to high-resolution remote sensing images. More importantly, all of the above methods ignore key attributes of  remote sensing images: their scales and orientation variations. Therefore, it is necessary to develop a model that can target these characteristics of remote sensing images and thus enhance the segmentation performance.

In this paper, we propose a novel local global class-aware network called LOGCAN++. It has two key components, the global class-aware (GCA) module and the local class-aware (LCA) module. The GCA module is designed with reference to previous work \cite{acfnet,ocrnet}, which reduces the background noise interference by extracting the global class centers for class-level context modeling. Meanwhile, the LCA module aims to alleviate the intra-class variance, which innovatively utilizes the local class centers as intermediate perceptual elements to indirectly correlate the pixels with their global class centers. In particular, during local class center extraction process, we introduce an affine transform block (ATB). It transforms the default local window into a target quadrilateral based on the affine transform function, thus can adapt the local windows’ size, shape, and location to diverse geospatial objects. This way, the local class centers generated by ATB are based on the object contents, thus coping well with the scale and orientation variations of remote sensing images. The main contributions of this paper are summarized as follows:
\begin{itemize}
\item We design a novel local-global class perception strategy. The intra-class variance in remote sensing images is well mitigated by indirectly associating pixels with global class centers by using local class centers as intermediate perceptual elements.
\item We introduce the affine transformation block for the first time to extract local class centers to adapt to geospatial objects with different sizes, shapes and orientations. This can cope well with the scale and orientation variations of remote sensing images.
\item We propose LOGCAN++, a semantic segmentation model customized to the characteristics of remote sensing images. Extensive experimental results on three remote sensing image semantic segmentation benchmarks show that LOGCAN++ outperforms other state-of-the-art methods and achieves a better balance between accuracy and efficiency.
\end{itemize}

This paper is an extension of our previous conference paper \cite{logcan}. This work extends the previous work in three aspects. 1) We introduce affine changes to improve the LCA module. It copes well with scale and orientation variations of remote sensing images by converting the default local window to a target quadrilateral to accommodate geospatial objects of different sizes, shapes, and orientations. 2) We extend the attention operation in the LCA module to be multi-head. so as to improve its ability to capture complex pixel-class-center relationships and to help avoid the overfitting problem that may result from a single attention mechanism. 3) We standardized the experimental setup and added sufficient experiments. We first added experiments on the larger LoveDA dataset. In addition, we provide more quantitative comparison results, qualitative analysis, and ablation experiments. We demonstrate the effectiveness of LOGCAN++ with more adequate experimental data.

\section{RELATED WORKS}

\subsection{General semantic segmentation}

\noindent Semantic segmentation is essentially an extension of image classification, from overall image analysis to pixel-by-pixel analysis. 
Traditional semantic segmentation methods usually conduct hand-crafted feature descriptors for feature extraction at each pixel point. However, designing feature descriptors is very time-consuming and the robustness of manual features is poor, requiring extensive expert a priori knowledge.
The Fully Convolutional Network (FCN)\cite{fcn} innovatively proposes an end-to-end approach that specifically handles classification tasks at the pixel level. However, the limited receptive field of convolutional networks becomes the performance bottleneck. To solve this problem, subsequent research focuses on context modeling, aiming to expand receptive fields and capture global information through spatial and relational context modeling, thereby improving feature representation. Representative works include PSPNet\cite{pspnet}, DeepLab series\cite{semantic}\cite{Deeplab}\cite{deeplabv2}\cite{deeplabv3+}, DenseASPP\cite{denseaspp}, DMNet\cite{dmnet}. 

To further enhance relational context modeling, recent studies \cite{danet,flanet, dmsanet} frequently built their models based on attention operations. Wang et al. \cite{nonlocal} uses self-attention in such a way that a single feature in any location is aware of all other positional features. Fu et al. \cite{danet} introduces dual attention mechanism to better capture the relationship between features and improve the performance of semantic segmentation. Huang et al. \cite{ccnet} calculates cross-attention between the rows and columns of the feature graph, so that each position can perceive the context information of other positions and improve the accuracy of semantic segmentation.

Compared with intensive relationship modeling on the entire feature map, some studies \cite{acfnet,ocrnet,ranet,cpnet,docnet} conduct context modeling from the perspective of categories, and improving intra-class compactness by calculating the similarity between pixels and class-level context and weighted aggregation of class-level context. Zhang et al. \cite{acfnet} first proposed the concept of class center to extract the global context from the perspective of class. Yuan et al. \cite{ocrnet} proposes object context representation, which enhances the representation of each pixel by calculating the relationship between each pixel and each target region, and weighting all target region representations into target context representations. Jin et al. \cite{isnet} proposes to enhance pixel representation by aggregating image-level and semantic-level context information respectively.

Transformer \cite{transformer} has recently gained popularity in the computer vision area \cite{beit,detr,setr,botnet,deit}. For example, ViT \cite{vit} divides each input image into blocks and converting them into vector sequences. However, it is less efficient in processing large-size images and is not suitable for multi-scale, intensive prediction tasks such as image segmentation. To address such issues, Swin Transformer \cite{swintransformer} introduces a windowing mechanism and multi-scale design to address these limitations. In the field of semantic segmentation, Transformer has been effectively applied by the ViT application of Segmenter \cite{segmenter} and the multi-scale location-free coding strategy of SegFormer \cite{segformer}. 
Recently, mask classification-based methods become popular \cite{maskformer,mpformer,faseg,pem}. MaskFormer\cite{maskformer} translates semantic segmentation into a more general mask classification task, and Mask2Former\cite{mask2former} further improves the efficiency and performance of this method by masking the attention mechanism. 

In summary, the above methods have achieved remarkable results on natural images. However, when applied to remote sensing images, they perform unsatisfactorily by ignoring the characteristics of remote sensing images with complex backgrounds, scale and orientation variations, and high intra-class variance. 
For example, recent general segmentation methods such as Segmenter \cite{segmenter} and DDP \cite{ddp}, which possesses a larger number of parameters and computational consumption, have not shown performance on remote sensing images that matches the model size (details can be found in Table \ref{tab:vai}, \ref{tab:pot} and \ref{tab:loveda}). This motivates us to propose LOGCAN++, whose network structure is designed to target these properties of remote sensing images.

\begin{figure*}[t]
	\centering \includegraphics[width=0.95\textwidth]
       {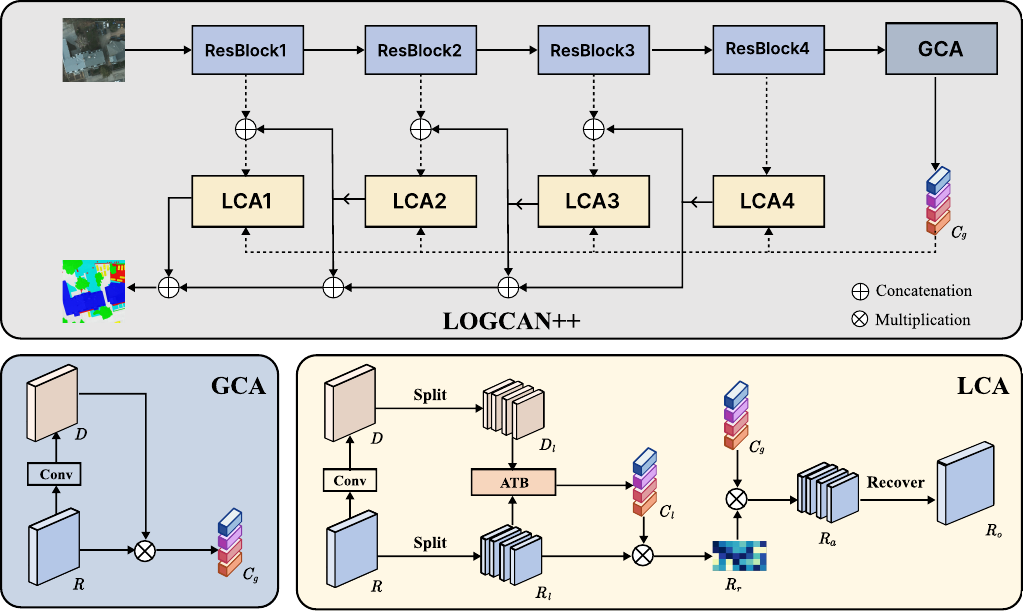}%
	\centering
	\caption{Architecture of the proposed LOGCAN++, which consists of backbone (ResNet-50 by default), local class aware (LCA) modules and a global class aware (GCA) module. GCA module generates global class centers for class level context modeling to reduce the background noise interference. LCA module generates local class centers based on affine transform block (ATB) and applies them as intermediate perceptual elements to indirectly correlate the pixels with the global class centers to mitigate the intra-class variance.} 
\label{fig:2}
\end{figure*}

\subsection{Semantic segmentation of remote sensing images}

\noindent Semantic segmentation approaches for remote sensing images can be categorized into two technical routes: application scenarios-specific approaches and application scenarios independent approaches.

The former approaches follow similar manners as general semantic segmentation approach with improvements for particular application scenarios as land use and land cover (LULC) classification \cite{land1}\cite{urban}\cite{jdp}, building extraction\cite{building1}\cite{lbe}\cite{building2}, road extraction\cite{bdtnet}\cite{bmda}\cite{roade2}\cite{roadtracer}, and vehicle detection\cite{vis}. For example, Jung et al.\cite{building1} introduce holistic edge detection (HED), extracting edge features on a structured encoder to enhance the clarity of building edges in remote sensing images. Bastani et al.\cite{roadtracer} enable the direct derivation of road network maps from the output of CNNs through an iterative search process guided by a CNN-based decision function. Mou et al.\cite{vis} effectively learn multi-level contextual feature representations from various residual blocks based on the residual learning principle. Luo et al.\cite{bdtnet} combine CNN and transformer technologies to design the bi-directional transformer module, adeptly capturing both global and local contextual information. However, these methods mainly focus on the improvement in special application scenarios and ignore the common problems in target segmentation of remote sensing images such as large intra-class variance, scale and orientation variations, and their applications are limited.

Recently, segmentation models for generic application scenarios (e.g., LANet\cite{lanet}, MANet \cite{manet}, PointFlow\cite{pointflow}, SCO\cite{sco}, GLOTs \cite{glots} and Farseg++\cite{farseg++}) also have  been widely investigated. However, these methods still fail to fully consider the characteristics of remote sensing images with complex backgrounds, scale and orientation variations, and high intra-class variance, limiting the further improvement of model performance. 
For example, Ding et al. \cite{lanet} proposes to enhance the embedding of contextual information using local attention blocks, which works to mitigate object scale variation and intra-class variance but ignores the challenges posed by complex backgrounds. Zheng et al. \cite{farseg,farseg++} enhances the differentiation of foreground features by learning foreground-scene relations to associate foreground-related contexts. 
Although it enhances the features in the foreground well, it ignores the large intraclass variance in remote sensing images, leading to poorer intraclass feature compactness. Especially, these methods have insufficient consideration for the characteristics of scale and orientation variations in remote sensing images. Therefore, the performance of these methods still needs to be improved due to the deficiencies in structural design.

\section{Methodology}

\subsection{Overall Architecture}

\noindent Since remote sensing images usually contain complex backgrounds, scale and orientation variations, and high intra-class variance, we propose a local-global class aware network LOGCAN++ to address these challenges. The overall architecture of the proposed LOGCAN++ is shown in Fig. \ref{fig:2}, which consists of a backbone, a set of local class aware (LCA) modules, and a global class aware (GCA) module. Specifically, the input high-resolution remote sensing image $I$ is first fed to the backbone to extract multi-scale features $R_1$, $R_2$, $R_3$, and $R_4$ (i.e., 1/4, 1/8, 1/16 and 1/32 of the original resolution, respectively.), where the deepest feature $R_4$ is input to the GCA module to obtain the global class center $C_g$. Then, we feed the feature $R_4$ along with the global class center $C_g$ to the LCA module for class-level context modeling. The context enhanced output features $R_o^4$ are concatenated with the shallow feature $R_3$, and input them into the LCA module again. This process is repeated three times to obtain the enhanced multi-scale features $R_o^1$, $R_o^2$, $R_o^3$ and $R_o^4$. Finally, the enhanced features are spliced along the channels and undergo a simple $1 \times1$ convolution to obtain the final prediction. The following sections provide the details of our GCA module, the LCA module and the loss function in detail.

\subsection{Global Class Aware Module}

\noindent There are a large number of irrelevant objects in the background of remote sensing images, which have different morphology and distribution, and large spectral differences.
Such extremely complex backgrounds would usually cause a large number of false alarms, which increase the difficulty of the scene parsing. The classical self-attention mechanism computes dense affinities between pixels over the entire image range, which introduces a large number of background pixels causing noise interference, and thus decrease the segmentation performance. Inspired by previous approaches \cite{ocrnet,acfnet} that proposing global class centers for class-level context modeling, we propose a global class-aware module.

Specifically, given the feature $R_4 \in \mathbb{R}^{C_4 \times H_4 \times W_4}$, we first obtain a pre-categorized representation $D_4 \in \mathbb{R}^{K \times H_4 \times W_4}$ by a simple $1 \times1$ convolution, where $K$ denotes the number of class. Then we obtain the global class center $C_g$ representing the average feature of the same class as:
\begin{equation}
    C_g = R_4 \otimes D_4^T.
\end{equation}
Here, we obtain the global class center $C_g$ based on the deepest feature $R_4$ only. This is due to the fact that the semantic information of the deepest feature is more accurate \cite{pspnet}, which is conducive to obtaining a more discriminative global class center. As a result, the obtained global class center $C_g$ can be used to enable class-level context modeling, in order to effectively avoids background noise interference during attention computation and enhancing the semantic representation of features. Please refer to previous work \cite{ocrnet,mdanet,logcan} for more details.

\begin{figure*}[t]
	\centering \includegraphics[width=0.9\textwidth]
       {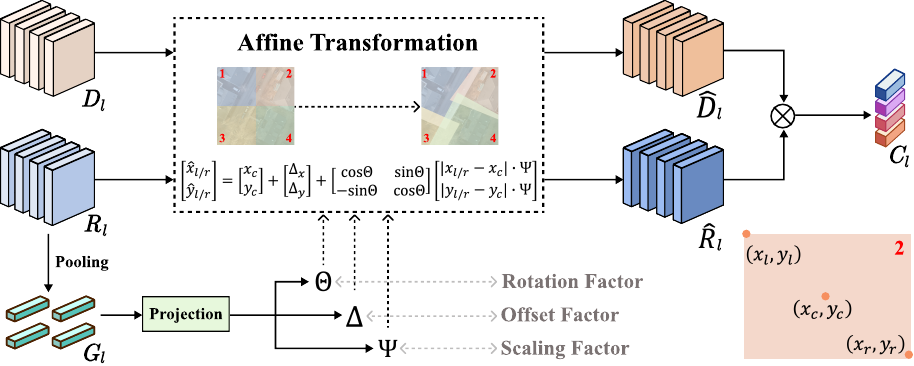}%
	\centering
	\caption{Structural details of the affine transform block (ATB). ATB first pools and projects features to produce scaling factors $\Psi$, offset factors $\Delta$, and rotation factors $\Theta$. These factors convert the default local window into a target quadrilateral to accommodate geospatial objects of different sizes, shapes, and orientations. As a result, the local class centers generated by ATB are data-conditional and thus cope well with scale and orientation variations in remote sensing images.}
\label{fig:3}
\end{figure*}

\subsection{Local Class Aware Module}

\noindent In remote sensing images, objects belonging to the same class, such as buildings, sometimes have large intra-class variance in the extracted semantic features due to differences in their sizes, shapes, colors, lighting conditions and imaging conditions. Such large intra-class variances leads to another problem for previous approaches \cite{ocrnet,acfnet} in their class-level context modeling: the semantic distance between features and global class centers is too far, affecting the accuracy of class-level context modeling. Considering the fact that objects of the same class in the local scope tend to be relatively similar, we propose a effective strategy which indirectly associates pixels with global class centers by using local class centers as intermediate perceptual elements.

For the feature representation $\mathcal{R}\in \mathbb{R} ^{C\times H\times W}$, we deploy a pre-classification operation for the corresponding distribution $\mathcal{D}\in \mathbb{R} ^{K\times H\times W}$. In particular, we split $\mathcal{R}$ and $\mathcal{D}$ along the spatial dimension to get $\mathcal{R}_l \in \mathbb{R}^{(N_h\times N_w) \times C\times ( h \times w)}$ and $\mathcal{D}_l\in \mathbb{R}^{(N_h\times N_w) \times K \times( h \times w)} $ as:
\begin{equation}
    R_l  = \text{Split}(R), D_l= \text{Split}(D),
\end{equation}
where $h$ and $w$ represent the height and width of the selected local patch, $N_h = \frac{H}{h}$, and $N_w = \frac{W}{w}$. 
Then, we introduce an affine transformation to convert the local window to the target quadrilateral to obtain $\hat{R}_l$ and $\hat{D}_l $. The local class center $C_l$ is then calculated as:
\begin{equation}
    C_l = \hat{R}_l \otimes \hat{D}_l^T.
\end{equation}
After the local class center $C_l$ is obtained, we implement the interaction between features and class centers using the multi-head cross-attention operation. Specifically, for each head $H_i$, the corresponding affinity matrix $R_r^i$, which represents the similarity between the pixel and the local class representations, is obtained as follows,
\begin{equation}
R_r^i = \text{Softmax}(\phi_r^i(R _{l})\otimes \phi_c^i(C_l)^T),
\end{equation} 
where $\phi_r$ and $\phi_c$ denote linear mapping operations. Finally, we utilize the obtained affinity matrix $R_r$ to associate the global class centers $C_g$ and acquire the augmented representations $R_a\in \mathbb{R}^{(N_h\times N_w) \times C\times ( h \times w)}$,
\begin{equation}
    R_a = \text{Concate}_{i=1}^{\epsilon}(R_r^i \otimes \phi_g(C_g)),
\end{equation}
where $\epsilon$ is the number of heads. Finally, we reduce the number of channels and recover the original spatial dimension of the augmented features to obtain the final feature $R_o$.
Here, we adopt the multi-head cross-attention to improve the model's ability of capturing complex pixel-class-centered relationships and avoid overfitting problems caused by the single attention mechanism. 
Consequently, the proposed LCA module captures more accurate pixel-class relationships by cleverly using the local class center as key and the global class center as value, which facilitates class-level context modeling. Please refer to Sec. \ref{stru} for the corresponding experimental analysis.

\subsection{Affine Transformation Block}

\noindent The ideal local class center should be the average semantic feature of objects in the local area. However, due to the Satellite sensor's unique overhead view and large imaging range, objects in remote sensing images usually present different orientations and possess significant scale differences. Fixed local windows often do not model these objects well. Therefore, we propose an affine transformation block (ATB) to extract higher quality local class centers, where the fixed local window is converted to a target quadrilateral to accommodate geospatial objects with different sizes, shapes, and orientations.

Specifically, the ATB block learns transformation factors based on the pixel feature content. Given local features $R_l$, it first obtains the local contextual features $G_l$ by average pooling as:
\begin{equation}
    G_l = \text{Pooling}(R_l).
\end{equation}
Then, we obtain the scaling factor $\Psi$ that resizes the local window, rotation factor $\Theta$ that orients the local window, and offset factor $\Delta$ that repositions the local window by feeding the obtained local contextual feature $G_l$ to linear mapping and LeakyReLU activation function as:
\begin{equation}
    \Theta, \Delta, \Psi = \text{LeakyReLU}(\phi_l(G_l)).
\end{equation}
where $\phi_l$ denotes the linear mapping operation that outputs three representations from the obtained $G_l$, which are undergo the LeakyReLU activation to generate the target scaling factor $\Psi$, rotation factor $\Theta$ and offset factor $\Delta$ of the corresponding local window. 
Then, we give a brief description of the affine transformation block. As shown in the lower right of Fig. \ref{fig:3}, given an initial local window, $x_l$, $y_l$ and $x_r$, $y_r$ denote the coordinates of the upper-left and lower-right corners of the initial window, respectively, while $x_c$, $y_c$ denote the coordinates of the window center. Then, its window transformation process is defined as:
\begin{equation}
\begin{bmatrix}\hat{x}_{l/r}\\\hat{y}_{l/r}\end{bmatrix}=\begin{bmatrix}x_{c}\\y_{c}\end{bmatrix}+\begin{bmatrix}\Delta_{x}\\\Delta_{y}\end{bmatrix}+\begin{bmatrix}\cos\Theta&\sin\Theta\\-\sin\Theta&\cos\Theta\end{bmatrix}\begin{bmatrix}|x_{l/r}-x_{c}|\cdot\Psi\\|y_{l/r}-y_{c}|\cdot\Psi\end{bmatrix},
\end{equation}
where $\hat{x}_{l}$, $\hat{y}_{l}$ and $\hat{x}_{r}$, $\hat{y}_{r}$ denote the coordinates of the upper-left and lower-right corners of the transformed window, respectively. $\Delta_{x}$ and $\Delta_{y}$  denote the offset in the horizontal and vertical directions, respectively. 
We then follow this formula to scale, rotate and offset the initial window to obtaind transformed local featurte $\hat{R}_l$ and corresponding pre-classified distribution $\hat{D}_l$. Since these factors are learned based on the pixel features, the transformed local window can be adaptively adjusted according to the size and orientation of the feature object. As a result, our ATB can help to cope with the scale and orientation changes in the remote sensing image well.

\subsection{Loss Function}

\noindent Due to the efficient design on the model structure, we only need the cross-entropy loss to supervise the training process of the proposed LOGCAN++. Specifically, the employed cross-entropy loss $\mathcal{L}_{ce}$ is defined as: 
\begin{equation}
    \mathcal{L}_{ce} = -\hat{p}\log(p)-(1-\hat{p})\log(1-p),
\end{equation}
where $\hat{p}$ and $p$ denote the ground-truth mask and the predict mask, respectively. Then, the overall losses for supervising the model training can be formulated as:
\begin{equation}
    \mathcal{L} = \lambda_{\text{main}}\mathcal{L}_{ce}^{main} + \lambda_{\text{aux}}\mathcal{L}_{ce}^{aux},
\end{equation}
where $\mathcal{L}_{ce}^{main}$ and $\mathcal{L}_{ce}^{aux}$ denote the cross-entropy loss applied to the final prediction and pre-classified representations $D$, respectively. $\lambda_{\text{main}}$ and $\lambda_{\text{aux}}$ denote the corresponding loss weights.

\begin{table*}[t]
\centering
\caption{Comparison with state-of-the-art methods on the test set of the ISPRS Vaihingen dataset. Per-class best performance is marked in bold. The FLOPs is calculated with the size of $512 \times 512 \times3$.}
\setlength{\tabcolsep}{6pt}{
\begin{tabular}{c| c| c c c c c c| c c c|c c}
\toprule
Method &Backbone &Imp. &Bui. & Low. &Tree & Car & Clutter &mIoU &mAcc &F1 &FLOPs(G) &Params(M)\\
\midrule
PSPNet \cite{pspnet} &ResNet-50 &\bf86.36 &91.69 &\bf72.65 &\bf80.87 &77.26 &32.32 &73.53 &80.23 &82.99 &178.45 &48.97\\
DeepLabV3+ \cite{deeplabv3+}&ResNet-50 &85.39 &91.55 &70.52 &79.49 &75.55 &35.22 &72.95 &80.27 &82.86 &176.25 &43.58\\
Semantic FPN \cite{fpn}&ResNet-50 &85.66 &91.19 &71.84 &80.08 &76.60 &33.05 &73.07 &79.98 &82.78 &\bf45.32 &\bf28.50\\
DANet \cite{danet}&ResNet-50 &84.80 &90.26 &69.79 &79.30 &74.32 &30.49 &71.50 &78.93 &81.55 &199.07 &49.82\\
OCRNet \cite{ocrnet}&HRNet-32 &85.61 &91.37 &69.84 &79.37 &77.77 &36.31 &73.38 &80.82 &83.21 &95.44 &32.94\\
FarSeg \cite{farseg}&ResNet-50 &85.96 &91.55 &71.95 &80.63 &78.24 &33.69 &73.67 &80.45 &83.20 &48.22 &31.39\\
Segmenter \cite{segmenter}&Swin-B &78.69 &82.17 &65.03 &75.85 &34.18 &15.03 &58.49 &67.12 &70.07 &69.62 &102.39\\
DC-Swin \cite{dcswin}&Swin-B &83.39 &87.76 &69.34 &78.89 &59.07 &18.42 &66.15 &73.69 &76.65 &124.32 &118.93\\
UnetFormer \cite{unetformer}&ResNet-50 &85.46 &91.16 &70.26 &79.46 &74.27 &35.44 &72.67 &80.00 &82.70 &57.20 &32.00\\
EfficientViT\cite{EfficientViT}&EfficientViT-L2 &83.87&88.39&70.49&79.69&61.28&15.06&66.46&74.25&76.44 &44.59 &51.32\\
DDP \cite{ddp}&Swin-T &85.73&91.42&71.63&80.32&78.32&33.52&73.49&80.54&83.07 &249.97 &39.56\\
RSSFormer \cite{10026298}  &RSS-B &86.28&91.25&72.49&80.94&74.00&35.21&73.36&81.00&83.12 &87.84 &32.14\\
\rowcolor{mygray}
LOGCAN++ &ResNet-50 &85.87 &\textbf{91.85} &71.44 &80.35 &\textbf{80.14} &\bf38.63 &\textbf{74.72} &\textbf{81.59} &\textbf{84.22} &51.26 &31.05\\
    \bottomrule	
    \end{tabular}
}
\label{tab:vai}
\end{table*}

\begin{figure*}[t]
	\centering \includegraphics[width=1.0\textwidth]
       {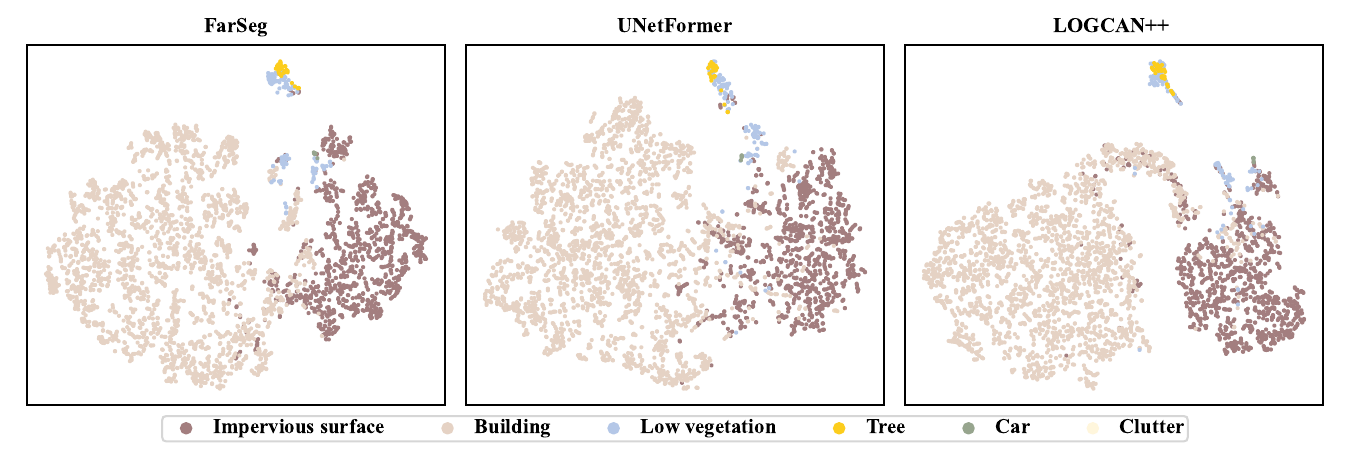}%
	\centering
	\caption{Visualization of features output by the last layer of FarSeg, UNetFormer and LOGCAN++. The test image is selected from ISPRS Vaihingen dataset. We implement the experiment with t-SNE \cite{tsne}.}
\label{fig:vai_tsne}
\end{figure*}

\section{Experimental settings}

\subsection{Datasets}

\noindent We evaluate our approach on three publicly available high-resolution RS datasets, including the ISPRS Vaihingen dataset \cite{www.isprs.org}, the ISPRS Potsdam dataset \cite{www.isprs.org}, and the LoveDA dataset \cite{loveda}.

\textbf{Vaihingen:} The Vaihingen dataset is released by the International Society for Photogrammetry and Remote Sensing (ISPRS) in 2014, and has become a widely-used benchmark dataset for high-resolution remote sensing image semantic segmentation. It includes high-resolution aerial images recorded in Vaihingen, Germany, with detailed manual annotations for six land cover classes: impervious surfaces, buildings, low vegetation, trees, cars, and clutter/background. The dataset contains 33 true orthophoto (TOP) images, each with a spatial resolution of 0.09 m/pixel, and widths ranging from 1887 to 3816 pixels. To evaluate segmentation performance, we used IDs 1, 3, 5, 7, 11, 13, 15, 17, 21, 23, 26, 28, 30, 32, 34, 37 for training, reserving the other 14 tiles for testing. The TOP images were segmented into 512 × 512 pixel patches with a 512-pixel stride, optimizing processing and analysis.

\textbf{Potsdam:} The Potsdam dataset also has been widely used for developing/evaluating high-resolution remote sensing semantic segmentation systems, which reflects the structure of the Vaihingen dataset. It consists of 38 high-quality image tiles, each with a ground sample distance (GSD) of 5 cm and a resolution of 6000 x 6000 pixels. This dataset includes multiple spectral bands (red, green, blue, near-infrared), a Digital Surface Model (DSM), and a Normalized Digital Surface Model (NDSM). In our experiments, we typically focus on three bands: red, green, and blue. Following the same division method as the Vaihingen dataset, researchers generally use tiles with IDs 2\_10, 2\_11, 2\_12, 3\_10, 3\_11, 3\_12, 4\_10, 4\_11, 4\_12, 5\_10, 5\_11, 5\_12, 6\_7, 6\_8, 6\_9, 6\_10, 6\_11, 6\_12, 7\_7, 7\_8, 7\_9, 7\_11, and 7\_12 for training, while reserving the remaining 14 tiles for testing.

\textbf{LoveDA:} The LoveDA dataset has been launched in 2021, which is a comprehensive dataset for high-resolution remote sensing image semantic segmentation, renowned for its complex backgrounds, diverse object scales, and varied class distributions. It encompasses 5987 images, where each contains 1024 × 1024 pixels and belongs to one of the seven classes: building, road, water, vegetation, car, clutter, and background. Covering an area of 536.15 km$^2$ with a ground sampling distance of 0.3 m, the dataset includes both urban and rural scenes from three Chinese cities - Nanjing, Changzhou, and Wuhan. It poses significant challenges due to multi-scale objects and inconsistent class distributions. The standard division includes 2522 images for training, 1669 images for validation, and 1796 images for testing.

\begin{figure*}[t]
	\centering \includegraphics[width=0.9\textwidth]
       {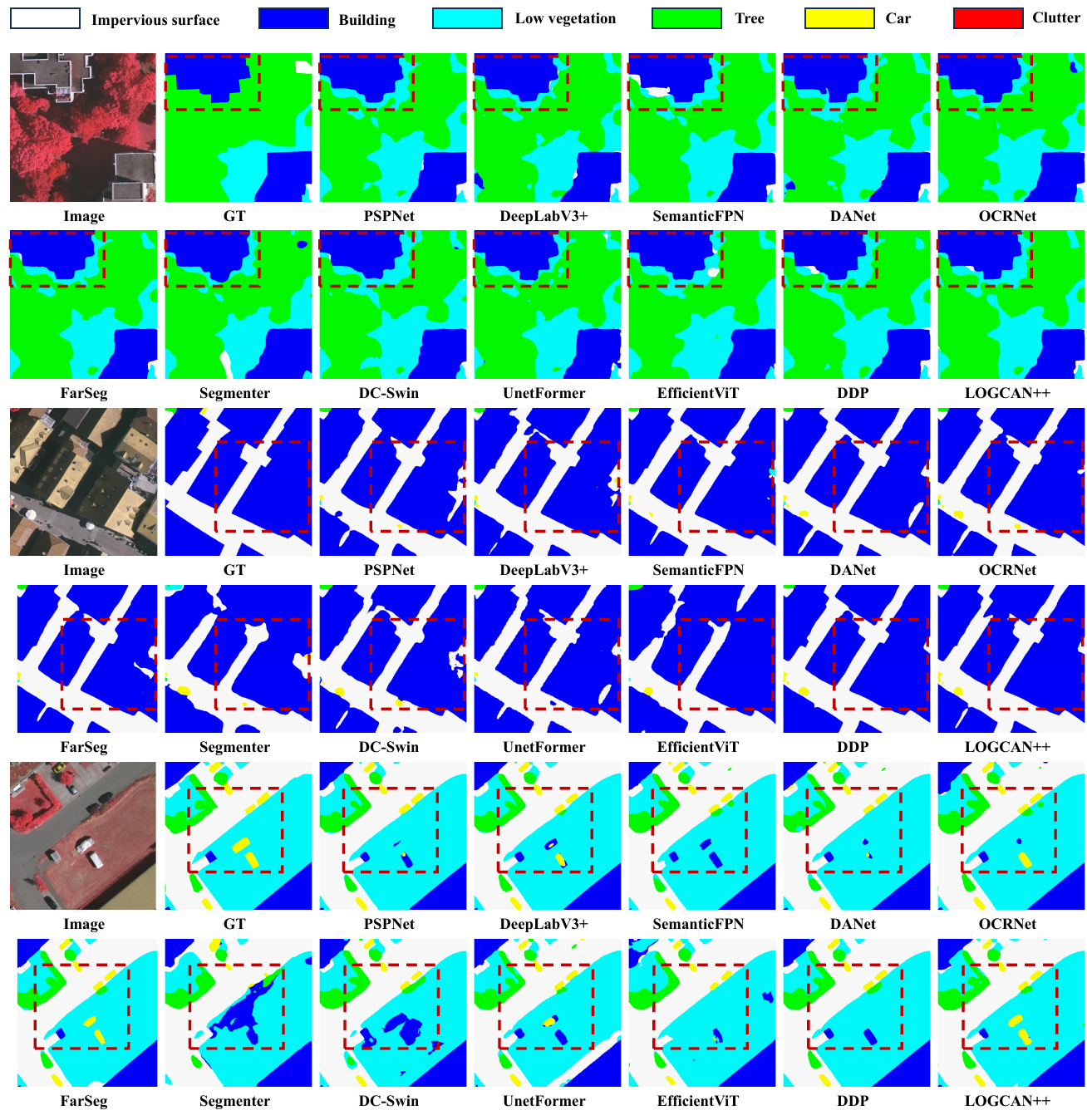}%
	\centering
	\caption{Qualitative comparison between LOGCAN++ and other state-of-the-art methods on the Vaihingen test set. The red dashed box is the area of focus. Best viewed in color and zoom in.}
\label{fig:vai}
\end{figure*}

\subsection{Evaluation Metrics}

\noindent In this paper, we employ two commonly used metrics to evaluate the performance of our approach: the mean Intersection over Union (mIoU), the mean accuracy (mAcc), and the F1 Score (F1).
\begin{itemize}
    \item \textbf{1) Mean Intersection Over Union (mIoU)}: The mIoU is a fundamental metric for assessing the performance of semantic segmentation models. It is calculated as the average Intersection over Union (IoU) across each class. Given that true positives (TP), false positives (FP), and false negatives (FN) represent the counts of pixels correctly classified, incorrectly classified, and missed, respectively, the mIoU is mathematically expressed as:
    \begin{equation}
    \text{mIoU} = \frac{1}{N} \sum_{i=1}^{N} \frac{\text{TP}_i}{\text{TP}_i + \text{FP}_i + \text{FN}_i}
    \end{equation}
    where \(N\) denotes the number of classes.

    \item \textbf{2) Mean Accuracy (mAcc)}: The mAcc is another pivotal metric for the evaluation of segmentation performance. It quantifies the average accuracy per class, defined by the ratio of true positives to the sum of true positives and false negatives. This is formulated as:
    \begin{equation}
    \text{mAcc} = \frac{1}{N} \sum_{i=1}^{N} \frac{\text{TP}_i}{\text{TP}_i + \text{FN}_i}
    \end{equation}

    \item \textbf{3) F1 Score (F1)}: The F1 Score is a critical metric for evaluating the balance between precision and recall in classification tasks. It is particularly useful in scenarios where an equitable importance is attributed to both false positives and false negatives. The F1 Score is the harmonic mean of precision and recall, thus providing a single metric to assess model performance when both the reduction of false positives and the increase of true positives are equally valued. The mathematical formulation of the F1 Score is given by:
    \begin{equation}
    \text{F1 Score} = 2 \times \frac{\text{Precision} \times \text{Recall}}{\text{Precision} + \text{Recall}}
    \end{equation}
    where \text{Precision} is defined as the ratio of true positives (\text{TP}) to the sum of true positives and false positives (\text{FP}), and \text{Recall} is the ratio of true positives to the sum of true positives and false negatives (\text{FN}).
    
\end{itemize}

\begin{table*}[t]
\centering
\caption{Comparison with state-of-the-art methods on the test set of the ISPRS Postdam dataset. Per-class best performance is marked in bold. The best value in each column is bolded.}
\setlength{\tabcolsep}{10pt}{
\begin{tabular}{l | c | c c c c c c | c c  c }
\toprule
Method &Backbone &Imp. &Building & Low. &Tree & Car & Clutter &mIoU &mAcc &F1\\
\midrule
PSPNet \cite{pspnet}&ResNet-50 &86.96 &93.59 &76.44 &79.06 &92.62 &38.55 &77.87 &84.68 &86.08\\ 
DeepLabV3+ \cite{deeplabv3+}&ResNet-50 &85.99 &92.44 &75.49 &77.84 &92.18 &40.06 &77.33 &84.99 &85.87\\
Semantic FPN \cite{fpn}&ResNet-50 &86.95 &93.63 &76.71 &\bf 79.85 &92.63 &34.15 &77.32 &83.94 &85.40\\
DANet \cite{danet}&ResNet-50 &85.92 &93.13 &74.62 &78.63 &91.91 &35.90 &76.68 &83.92 &85.16\\
OCRNet \cite{ocrnet}&HRNet-32 &86.01 &91.66 &75.91 &78.49 &91.00 &\bf 40.66 &77.29 &85.00 &85.91\\ 
FarSeg \cite{farseg}&ResNet-50 &86.93 &93.56 &76.72 &79.00 &91.37 &40.32 &77.98 &84.90 &86.29\\
Segmenter \cite{segmenter}&Swin-B &80.54 &86.12 &70.64 &68.35 &77.34 &26.60 &68.27 &77.91 &79.17\\ 
DC-Swin \cite{dcswin}&Swin-B &84.88 &90.47 &74.95 &76.42 &89.16 &34.83 &75.12 &82.80 &84.18\\ 
UnetFormer \cite{unetformer}&ResNet-50 &86.50 &93.33 &75.80 &78.77 &92.43 &38.06 &77.48 &84.48 &85.81\\
EfficientViT\cite{EfficientViT}&EfficientViT-L2 &83.58&89.38&73.91&73.77&88.74&30.92&73.38&81.25&82.77\\
DDP \cite{ddp}&Swin-T &86.88&93.5&75.63&78.79&92.59&39.24&77.77&84.78&86.07\\
RSSFormer \cite{10026298} &RSS-B &86.92&93.12&77.19&79.55&91.43&38.88&77.85&84.55&86.01\\
\rowcolor{mygray}
LOGCAN++ &ResNet-50 &\textbf{87.51} &\textbf{93.76} &\textbf{77.20} &79.78 &\textbf{93.13} &40.10 &\textbf{78.58} &\textbf{85.34} &\textbf{86.62}\\
    \bottomrule	
    \end{tabular}
}
\label{tab:pot}
\end{table*}

\begin{figure*}[t]
	\centering \includegraphics[width=1.0\textwidth]
       {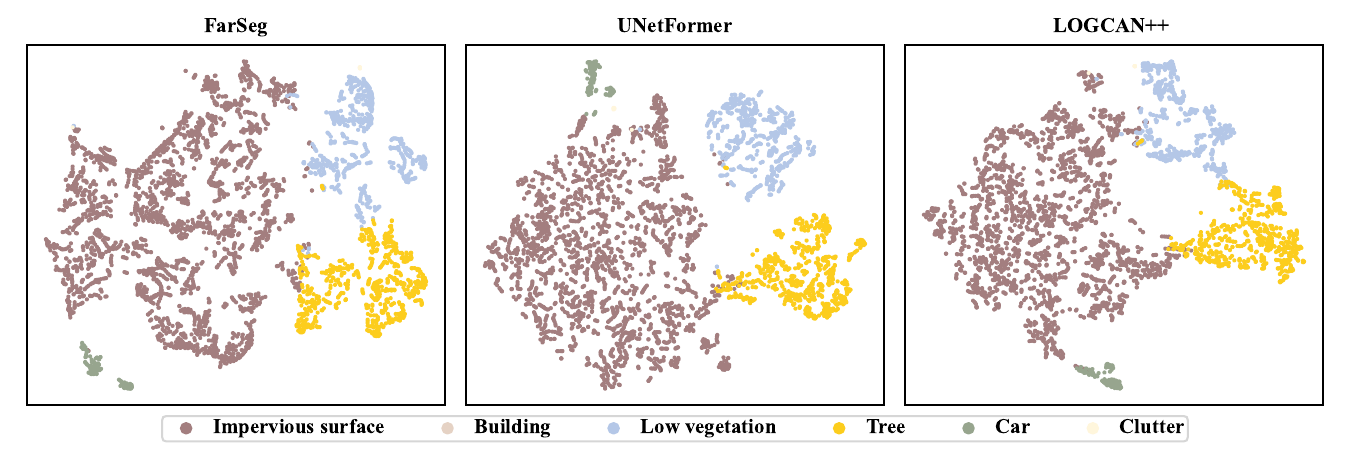}%
	\centering
	\caption{Visualization of features output by the last layer of FarSeg, UNetFormer and LOGCAN++. The test image is selected from ISPRS Potsdam dataset. We implement the experiment with t-SNE \cite{tsne}.}
\label{fig:pot_tsne}
\end{figure*}

\subsection{Implementation Details}

\noindent We choose ResNet-50 \cite{resnet} as the backbone for a fair comparison \cite{farseg++, lanet}. For our training setup, we utilize a 512 × 512 pixel input image resolution with a batch size of 8. The optimization is managed using the Stochastic Gradient Descent (SGD) optimizer, with an initial learning rate of 0.01, momentum of 0.9, and a weight decay of 0.0005. The training, comprising 80,000 iterations, incorporates a polynomial learning rate strategy. Data augmentation techniques such as random flipping, scaling (with a scaling range of 0.5 to 1.5), and photometric distortion are employed to boost the model's generalization ability. Testing includes multiscale enhancement implemented involve image scaling within a range of 0.5 to 1.5 and random flipping. All our scripts are implemented using the PyTorch framework \footnote{\url{https://pytorch.org/}} on NVIDIA A6000 GPUs with 48 GB of memory, operating under a Linux Ubuntu 20.04 system.

\begin{figure*}[t]
	\centering \includegraphics[width=0.9\textwidth]
       {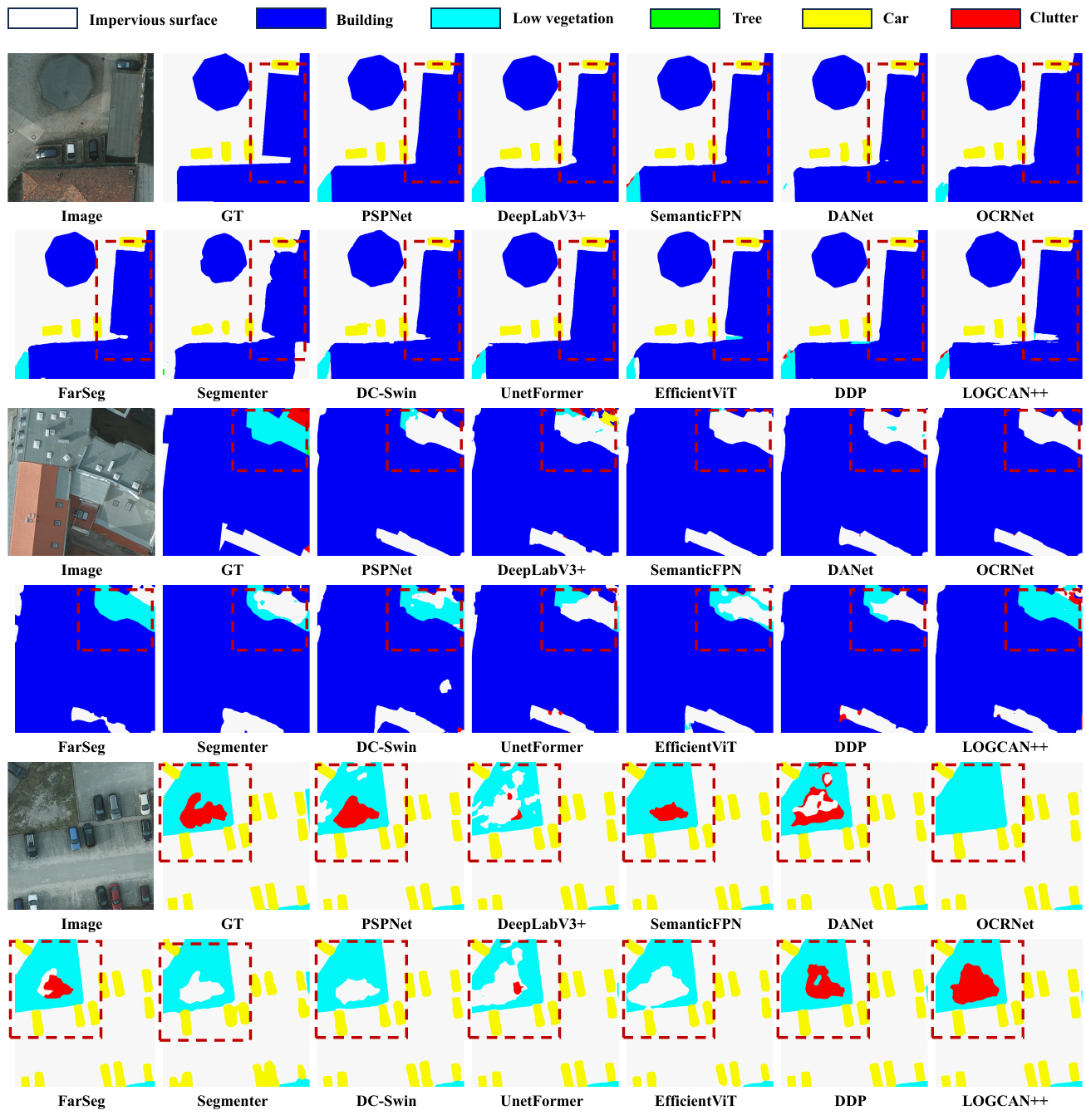}%
	\centering
	\caption{Qualitative comparison between LOGCAN++ and other state-of-the-art methods on the Potsdam test set. The red dashed box is the area of focus. Best viewed in color and zoom in.}
\label{fig:pot}
\end{figure*}

\section{EXPERIMENTAL RESULTS AND ANALYSIS}

\subsection{Results on the ISPRS Vaihingen dataset}
\subsubsection{Quantitative analysis}
To validate the effectiveness of our LOGCAN++, we first carry out experiments on the Vaihingen dataset. The comparison methods include convolution-based methods such as PSPNet \cite{pspnet}, DeepLabV3+ \cite{deeplabv3+}, Semantic FPN \cite{fpn}, DANet \cite{danet}, OCRNet \cite{ocrnet}, FarSeg \cite{farseg}, transformer-based methods such as segmenter \cite{segmenter}, DC-Swin \cite{dcswin}, UNetFormer \cite{unetformer},  EfficientViT \cite{EfficientViT} and diffusion based methods such as DDP \cite{ddp}. As shown in Table \ref{tab:vai}, our LOGCAN++ achieved the optimal segmentation performance. Specifically, compared to recent general segmentation methods such as DDP and RS segmentation methods such as UNetFormer, we have an increase of 2.05\% and 1.23\% in mIoU, respectively. In particular, for classes with large variance such as buildings, our approach achieved a mIoU value of 91.85\%, which outperforms other mainstream methods, validating the effectiveness of our local global class aware strategy. In addition, for small objects such as car, we have a 1.82\% mIoU improvement compared to DDP. This indicates that the introduction of affine transformations enhanced the fine-grained recognition of small objects. 

In addition, LOGCAN++ possesses good efficiency compared to a series of recent popular works. Specifically, we first remove the GCA module and replace it with local class centers, resulting in a decrease of 0.1M in model parameters and 0.01G in FLOPs. Next, we eliminate the LCA module and substitute the local class centers with global class centers, which leads to a reduction of 2.06M in model parameters and 10.744G in FLOPs. These results demonstrate the lightweight of the LCA and GCA module designs. When compared to the recent state-of-the-art work DDP, LOGCAN++ requires only 78\%  parameters and 21\% FLOPs of DDP, but exhibits superior performance (i.e., 84.22 in F1). Compared to the lightweight methods UNetFormer and EfficientViT, LOGCAN++ has similar or slightly better efficiency, yet achieves a significant performance improvement, with 1.52 and 7.78 improvements in F1 metrics, respectively. Overall, the proposed LOGCAN++ achieve a better trade-off between accuracy and efficiency.

\subsubsection{Qualitative analysis}
To explore the impact of our LOGCAN++ on the pixel feature generation process, Fig. \ref{fig:vai_tsne} visualizes the feature distribution on images from the vaihingen test set. Specifically, we visualize the features output by the last layer of FarSeg, LOGCAN++ and UNetFormer using t-SNE \cite{tsne}. It can be observed that the feature distributions extracted by FarSeg and UNetFormer for the impervious surfaces and buildings categories cross each other, whereas the features extracted by LOGCAN++ for these two categories show good separation. Moreover, for the building category, the features extracted by LOGCAN++ are more compact. This can be attributed to the fact that our local global class aware strategy mitigates the intra-class variance well and achieves a better interaction with the global class centers, thus extracting more discriminative features. In conclusion, LOGCAN++ extracts features with higher intra-class compactness and inter-class separation than FarSeg and UNetFormer.

In addition, Fig. \ref{fig:vai} visualizes the segmentation masks output by different models to qualitatively compare the segmentation performance of our LOGCAN++ with these competitors, where all input images come from the Vaihingen test set. 
Specifically, for the first input image, UNetFormer \cite{unetformer} and EfficientViT \cite{EfficientViT} suffer from mask boundary blurring on the building class. For the second input image, FarSeg \cite{farseg} suffers from fragmentation in the building class mask, i.e., some pixels are misclassified as impervious surfaces. In contrast to these works, buildings have more complete shapes and clearer boundaries for both images in the output mask of LOGCAN++.
For the third image, our LOGCAN++ is able to recognize cars accurately, while other methods such as UNetFormer and DDP \cite{ddp} misidentify them as buildings. 
In addition, LOGCAN++ provides a complete segmentation of low-vegetation areas, whereas FarSeg appears to have some pixels that are misclassified as impervious surfaces due to these pixels having very similar spectra and shapes with impervious surfaces.
These results show that our LOGCAN++ has better semantic recognition ability and visualization performance.

\begin{table*}[t]
\centering
\caption{omparison with state-of-the-art methods on the test set of the LoveDA dataset. Per-class best performance is marked in bold. Please note that the LoveDA dataset requires an online test to evaluate the model. Therefore results for the F1 and mAcc metrics are not available here. The best value in each column is bolded.}
\setlength{\tabcolsep}{10pt}{
\begin{tabular}{l|c|ccccccc|c}
\toprule
Method &Backbone &Background &Building & Road &Water &Barren &Forest &Agriculture &mIoU\\
\midrule
PSPNet \cite{pspnet}&ResNet-50 &45.81 &58.21 &\bf58.37 &80.23 &14.19 &46.55 &62.79 &52.31\\
DeepLabV3+ \cite{deeplabv3+}&ResNet-50 &45.31 &55.92 &57.09 &78.33 &11.82 &45.87 &63.89 &51.18\\
Semantic FPN \cite{fpn}&ResNet-50 &45.45 &56.84 &57.54 &79.43 &16.78 &45.89 &63.43 &52.19\\
DANet \cite{danet}&ResNet-50 &46.44 &56.84 &57.03 &79.31 &14.65 &46.59 &64.66 &52.22\\
OCRNet \cite{ocrnet}&ResNet-50 &46.05 &58.66 &57.17 &\bf80.72 &13.27 &46.83 &64.53 &52.46\\
FarSeg \cite{farseg}&ResNet-50 &45.18 &56.55 &56.92 &79.78 &14.96 &46.59 &62.50 &51.78\\
Segmenter \cite{segmenter}&Swin-B &39.7 &44.71 &47.29 &73.51 &8.35 &37.45 &55.71 &43.82\\
DC-Swin \cite{dcswin}&Swin-B &42.34 &50.17 &54.83 &77.13 &8.16 &42.76 &59.28 &47.81\\
UnetFormer \cite{unetformer}&ResNet-50 &44.16 &56.51 &55.84 &79.00 & 13.86 & 44.27 &62.65 &50.90\\
EfficientViT \cite{EfficientViT}&EfficientViT-L2 &42.92 &50.98 &52.78 &75.65 &4.27 &42.04 &61.19 &47.12\\
DDP \cite{ddp}&Vit-B &46.19 &57.20 &58.22 &80.34 &14.93 &46.54 &64.34 &52.54\\
RSSFormer \cite{10026298} &RSS-B &44.03&57.23&55.35&79.37&10.91&45.22&62.04&50.59\\
\rowcolor{mygray}
LOGCAN++ &ResNet-50 &\textbf{47.37} &\bf58.38 &56.46 &80.05 &\textbf{18.44} &\textbf{47.91} &\textbf{64.80} &\textbf{53.35}\\

    \bottomrule	
    \end{tabular}
}
\label{tab:loveda}
\end{table*}

\subsection{Results on the ISPRS Potsdam dataset}
\subsubsection{Quantitative analysis}
We also conduct experiments on the Potsdam dataset to validate the effectiveness of LOGCAN++. As shown in Table \ref{tab:pot}, our LOGCAN++ outperformed all other state-of-the-art methods with clearly advantages. Specifically, compared to recent segmentation methods such as EfficientViT and DDP, our approach gained an improvement of 5.20\% and 0.81\% on the mIoU metric, respectively. In particular, it exhibits a similar phenomenon on the Potsdam dataset as on the Vaihingen dataset: for classes with large variance such as buildings and impervious surfaces, it achieved mIoU values of 93.76\% and 87.51\%, respectively. Meanwhile, for small objects such as car, it achieved mIoU of 93.13\%, which largely outperforms other methods. This further verifies that LOGCAN++ is able to cope well with the challenges posed by the remote sensing images due to the characteristics of complex backgrounds, scale and orientation variations, and large intra-class variance through the local global class awre strategy and affine transform.

\subsubsection{Qualitative analysis}
Similarly, Fig. \ref{fig:pot_tsne} visualizes the feature distribution for images from the Potsdam test set. For categories such as impervious surfaces and trees, LOGCAN++ exhibits better intra-class compactness. We argue that the reason for this is that impervious surfaces and trees in the aerial view tend to have irregular shapes and frequent changes in scale and orientation. Our LOGCAN++ is able to mitigate the problem of large feature variance within-class caused by the above phenomena based on the affine transformations. In addition, LOGCAN++ has less crossover between feature distributions of different classes. The qualitative results indicate that LOGCAN++ is able to acquire higher quality semantic features. In addition, we visualize the segmentation masks output by our models and other competitors. As illustrated in Fig. \ref{fig:pot}, the first image shows that LOGCAN++ is able to preserve clearer building boundaries; the second image shows that LOGCAN++ is able to recognize low vegetation in complex backgrounds; and the third image shows that LOGCAN++ significantly reduces false alarms caused by the conplex background. These results demonstrate that our LOGCAN++ is able to significantly improve the quality of the segmentation mask compared to other methods due to the effectiveness of its structural design.

\newcommand{\up}[2]{
#1 \fontsize{6pt}{1em}\selectfont\color{mygreen}{$\!\uparrow\!$ \textbf{{#2}}}
}

\newcommand{\down}[2]{
#1 \fontsize{6pt}{1em}\selectfont\color{myblue}{$\!\downarrow\!$ \textbf{{#2}}}
}

\subsection{Results on the the LoveDA dataset}
\subsubsection{Quantitative analysis} The experimental results on the LoveDA dataset are compared in Table \ref{tab:loveda}. It is clear that our LOGCAN++ largely outperformed other state-of-the-art methods on the Loveda dataset. Compared to recent segmentation methods such as Efficientvit and DDP, our approach has a mIoU improvement of 6.23\% and 0.81\%, respectively. In addition, LOGCAN++ improves especially on confusing classes such as buildings and barrens. Note that compared to the vaihingen and potsdam datasets, the LoveDA dataset has a more complex context, more samples, and significantly different geographies (i.e., urban vs. rural) \cite{loveda}. Therefore, it is reasonable to conclude that the effective design allows our LOGCAN++ consistently achieved superior performances in more challenging segmentation scenarios.

\begin{figure*}[t]
	\centering \includegraphics[width=0.9\textwidth]
       {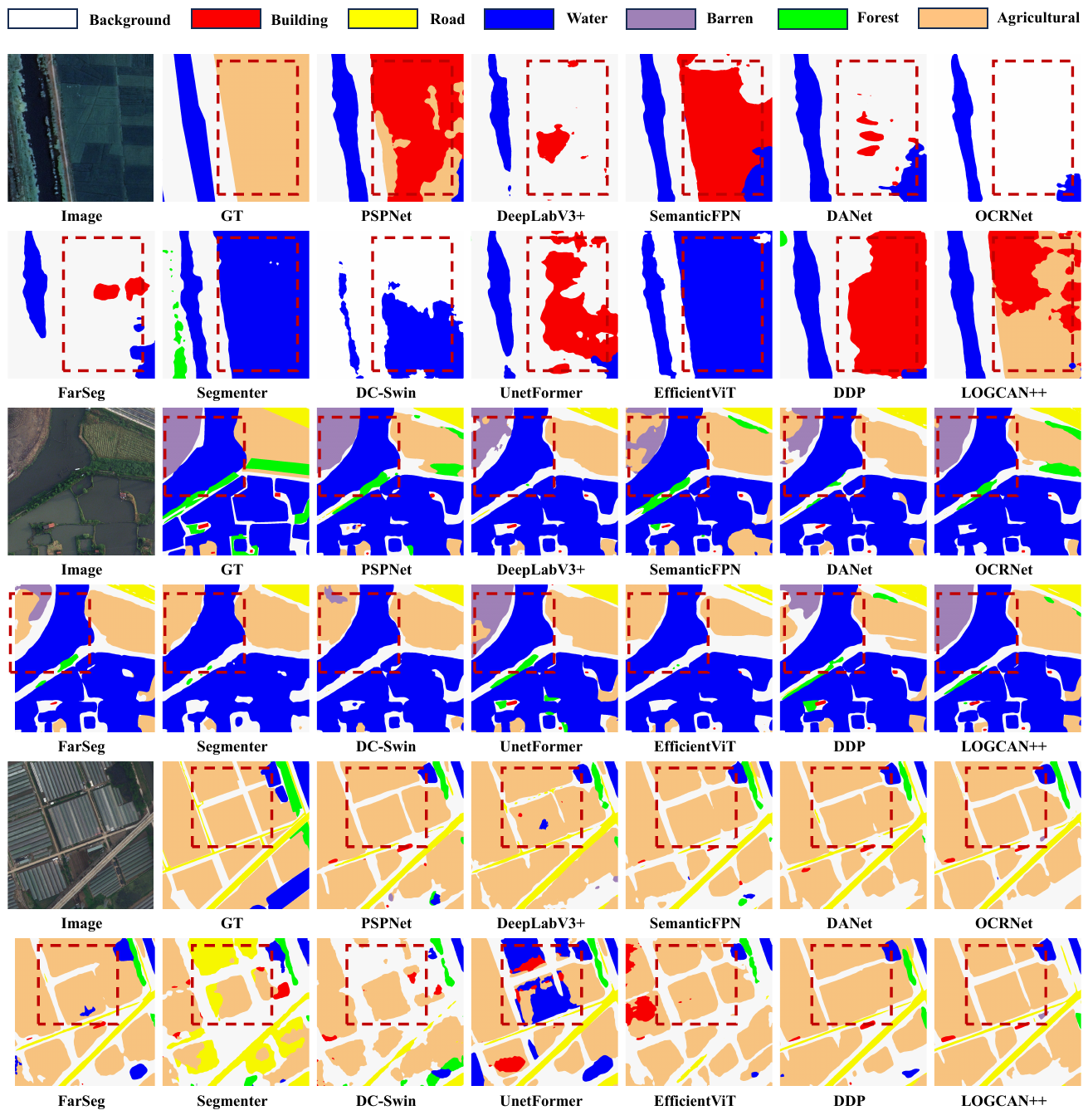}%
	\centering
	\caption{Qualitative comparison between LOGCAN++ and other state-of-the-art methods on the LoveDA test set. The red dashed box is the area of focus. Best viewed in color and zoom in.}
\label{fig:loveda}
\end{figure*}

\subsubsection{Qualitative analysis}
Fig. \ref{fig:loveda} visualizes segmentation masks for example images from the LoveDA validation set.
For the agriculture category in the first image, other comparison methods have misidentified it as building or water, while LOGCAN++ can recognize the correct semantics. Although its mask shape is not complete, our approach has shown better category recognition ability, which can be attributed to the fact that LOGCAN++ is able to enhance the discrimination of the features based on the local global class aware strategy. In addition, for the second and third images, LOGCAN++ has segmented more complete shapes, whereas the comparison methods tend to suffer from fragmentation of the masks. Such good visualization performances further validates the effectiveness of our LOGCAN in the task of remote sensing image semantic segmentation.

\subsection{Ablation study}
\label{stru}

\noindent To further validate the model design, we conduct a series of ablation experiments on the Vaihingen dataset. In summary, we perform ablation analysis for the model structure, the design of the affine transformation block (ATB), the number of local patches and the number of heads in the multi-head attention. These experiments and analyses motivate us to obtain the optimal model design. We describe it in detail next.

\subsubsection{Analysis of the model structure}
We first perform ablation analysis of the LOGCAN++ structure on the vaihingen dataset, as shown in Table \ref{tab:struct}. Specifically, we set up a series of variants: 1) both LCA and GCA modules are present, i.e., the final LOGCAN++ structure; 2) the GCA module is removed, i.e., the pixel features at each spatial scale are contextualized with the local class center; 3) the LCA module is removed, i.e., the pixel features at each spatial scale are contextualized with the global class center; and 4) both the GCA module and the LCA module are both removed, i.e., pixel features at each spatial scale are directly spliced. The results suggest that adding either the LCA module or the GCA module yields a miou improvement of 3.40\% and 3.07\%, respectively, while the combination of the two further improves the performance. This validates the effectiveness of the local global class perception strategy.

\begin{figure*}[t]
	\centering \includegraphics[width=0.9\textwidth]
       {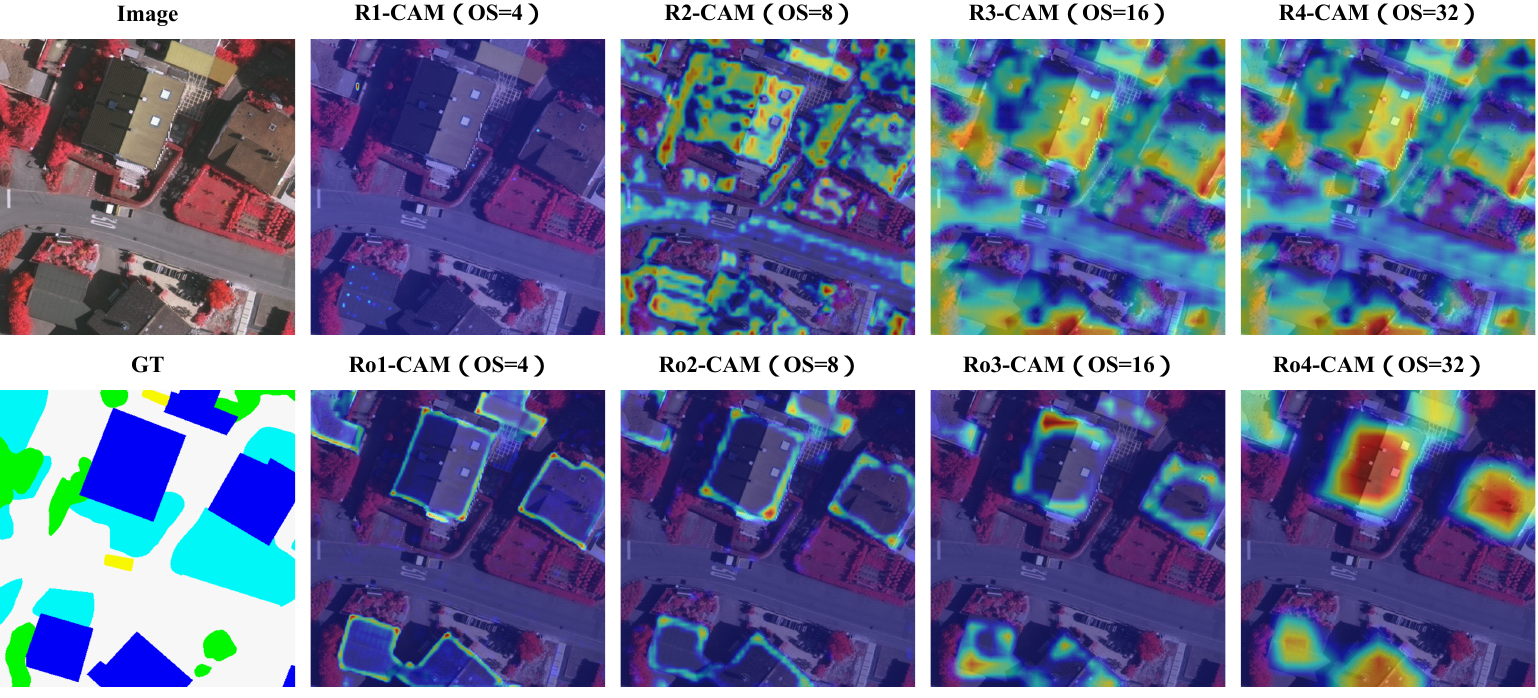}%
	\centering
	\caption{Class activation maps for the building category, which is based on grad-cam \cite{gradcam}. The input image is selected from the test set of the Vaihingen dataset. We analyze the features of LOGCAN++ at four scales to further help understand the working mechanism of LOGCAN++.}
\label{fig:cam}
\vspace{2mm}
\end{figure*}

\begin{table*}[t]
\centering
\caption{{Ablation Study on the Impact of Model Structure Variations on the Vaihingen Dataset. The first line is base. Green and blue numbers represent increases and decreases compared to base, respectively. The best value in each column is bolded.}}
\setlength{\tabcolsep}{8pt}{
\begin{tabular}{cc|c c c c c c|c}
\toprule
GCA & LCA &Impervious surface &Building & Low vegetation &Tree & Car & Clutter &mIoU\\
\midrule
\CheckmarkBold & \CheckmarkBold &85.87 &91.85 &71.44 &80.35 &80.14 &38.63 &74.72\\

\XSolidBrush & \CheckmarkBold &\down{85.81}{0.06} &\down{91.62}{0.23} &\up{71.81}{0.37} &\up{80.61}{0.26} &\down{77.92}{2.22} &\down{34.73}{3.90} &\down{73.75}{0.97}\\

\CheckmarkBold & \XSolidBrush &\up{86.03}{0.16}&\down{91.67}{0.18} &\up{71.59}{0.15}&\down{80.28}{0.07}&\down{79.10}{1.04}&\down{35.81}{2.82}&\down{74.08}{0.64}\\

\XSolidBrush & \XSolidBrush &\down{83.26}{2.61} &\down{88.69}{3.16} &\down{69.42}{2.02} &\down{77.64}{2.71} &\down{75.14}{5.00} &\down{29.93}{8.70} &\down{70.68}{4.04}\\
    \bottomrule	
    \end{tabular}
}
\label{tab:struct}
\end{table*}

\begin{table*}[t]
\centering
\caption{Ablation Study on the Impact of Using Different Layer Features for calculating Global Class Center \( C_g \) on the Vaihingen Dataset. The resolution increases from top to bottom. Green and blue numbers represent increases and decreases compared to base, respectively.}
\setlength{\tabcolsep}{8pt}{
\begin{tabular}{c|c c c c c c|c}
\toprule
generation of $C_g$ &Impervious surface &Building & low vegetation &Tree & Car & Clutter &mIoU\\
\midrule
$feature_4$ &85.87 &91.85 &71.44 &80.35 &80.14 &38.63 &74.72\\
$feature_3$  &\down{85.74}{0.07} &\down{91.18}{0.67} &\down{71.34}{0.10} &\up{80.38}{0.03} &\down{77.78}{2.36} &\down{34.33}{4.30} &\down{73.46}{1.26}\\
$feature_2$ &\down{85.73}{0.14} &\down{90.98}{0.87} &\up{71.33}{0.11} &\down{80.90}{0.55}&\down{76.72}{3.42} &\down{34.44}{4.19} &\down{73.35}{1.37}\\
$feature_1$ &\down{85.83}{0.04} &\down{90.97}{0.88} &\up{71.63}{0.19} &\down{79.67}{0.68} &\down{77.47}{2.67} &\up{39.08}{0.45} &\down{74.11}{0.62}\\
    \bottomrule	
    \end{tabular}
}
\vspace{1mm}
\label{tab:cg}
\end{table*}

To further understand how the GCA and LCA modules guide the learning of the model, we select an example image from the Vaihingen test set for visualization and analysis. As shown in Fig. \ref{fig:cam}, we set buildings as the target class to obtain the corresponding class activation map based on Grad-CAM \cite{gradcam}. It can be observed that the activation of the backbone in the building region gradually increases with the increasing of the model depth, which indicates that each layer in the model can learn unique and task-related cues, i.e., all layers of the model can learn task-specific and complementary cues. However, there are still a large number of activations occurring outside the building area. When guided by the GCA module and LCA module, the activations are more accurate and concentrated in the building area. In particular, for the shallow enhanced feature $R_o^1$, LOGCAN++ obtained higher activations than $R_1$ in the edge region of the building, which is key to the model's ability to segment clear building boundaries. Qualitative ablation analysis proves the effectiveness of LOGCAN++.

\subsubsection{Analysis of the generation of the global class center $C_g$}
Table \ref{tab:cg} compares the model performance achieved by applying four feature maps (of different resolutions) generated by four different layers of the backbone network to perform pre-classification and derive global class centers. Our findings reveal that when the model uses the feature map of the deepest layer, the model performance reaches its peak, which particularly show large performance improvements for small objects such as cars. We argue that the deepest feature map effectively eliminates much of the noise interference, allowing the model to focus on the critical information of small target objects and thereby enhancing its recognition capability.

\begin{table*}[t]
\centering
\caption{Ablation Study on the Impact of Different ATB Strategies on the Vaihingen Dataset. The first line is base. Green and blue numbers represent increases and decreases compared to base, respectively. }
\setlength{\tabcolsep}{8pt}{
\begin{tabular}{ccc|c c c c c c|c}
\toprule
$\Psi$ & $\Theta$ & $\Delta$ &Impervious surface &Building & low vegetation &Tree & Car & Clutter &mIoU\\
\midrule
\CheckmarkBold & \CheckmarkBold & \CheckmarkBold  &85.87 &91.85 &71.44 &80.35 &80.14 &38.63 &74.72\\
\XSolidBrush& \CheckmarkBold & \CheckmarkBold &\down{85.81}{0.06} &\down{91.46}{0.39} &\up{71.63}{0.19} &\down{80.33}{0.02}&\down{78.72}{1.42} &\down{33.95}{4.68} &\down{73.65}{1.07}\\
\CheckmarkBold & \XSolidBrush & \CheckmarkBold &\up{85.96}{0.09} &\down{91.65}{0.20} &\down{71.37}{0.07} &\down{80.16}{0.19} &\down{78.98}{1.16} &\down{37.13}{1.50} &\down{74.21}{0.51}\\
\CheckmarkBold & \CheckmarkBold & \XSolidBrush&\up{85.90}{0.03} &\down{91.57}{0.28} &\up{72.10}{0.66} &\up{80.57}{0.22} &\down{77.66}{2.48} &\down{36.38}{2.25} &\down{74.03}{0.69}\\

\XSolidBrush & \XSolidBrush & \XSolidBrush & \down{85.11}{0.76} &\down{91.13}{0.72} &\up{70.79}{0.65} &\up{79.92}{0.43} &\down{77.28}{2.86} &\down{33.99}{4.64} &\down{73.02}{1.70}\\
    \bottomrule	
    \end{tabular}
\vspace{4mm}
}
\label{tab:atb}
\end{table*}

\begin{table*}[t]
\centering
\caption{Ablation Study on the Impact of Patch Numbers (i.e., $N_h \times N_w$) on the Vaihingen Dataset. The highest value is bolded.}
\setlength{\tabcolsep}{10pt}{
\begin{tabular}{c|c c c c c c|c}
\toprule
$N_h \times N_w$ &Impervious surface &Building & low vegetation &Tree & Car & Clutter &mIoU\\
\midrule
$1\times1$&\up{86.03}{0.16}&\down{91.67}{0.18}&\up{71.59}{0.15}&\down{80.28}{0.07}&\down{79.10}{1.04}&\down{35.81}{2.82}&\down{74.08}{0.64}\\
$2\times2$ &\up{86.12}{0.25}&\down{91.50}{0.35}&\up{72.08}{0.64}&\up{80.57}{0.22}&\down{78.36}{1.78}&\down{33.44}{5.19}&\down{73.68}{1.04}\\
$4\times4$ &85.87 &91.85 &71.44 &80.35 &80.14 &38.63 &74.72\\
$8\times8$ &\down{84.67}{1.20}&\down{90.59}{1.26}&\down{69.55}{1.89}&\down{79.00}{1.35}&\down{76.66}{3.48}&\down{37.58}{1.05}&\down{73.01}{1.71}\\
$16\times16$ &\down{85.07}{0.80}&\down{91.46}{0.39}&\down{69.82}{1.62}&\down{78.97}{1.38}&\down{75.99}{4.15}&\down{33.46}{5.17}&\down{72.46}{2.26}\\
    \bottomrule	
    \end{tabular}
}
\vspace{4mm}
\label{tab:num}
\end{table*}

\subsubsection{Analysis of the ATB design}
We also conduct experiments on the Vaihingen dataset to verify the effectiveness of the ATB design. Specifically, we use the original ATB as the baseline and remove the scaling factor $\Psi$, rotation factor $\Theta$ and offset factor $\Delta$, respectively. The experimental results are shown in Table \ref{tab:atb}. When any one of these factors is removed, the performance of the model is degraded to some extent. In particular, when the scaling factor $\Psi$ is removed, the mIoU metric of the model decreases significantly. This can be explained by the fact that remote sensing images tend to have large scale variations, and a fixed local window cannot adapt to target objects of different scales. In addition, when the rotation and offset factors are removed, our model's ability to model objects with different orientations and positions decreases, which lead to the quality of the extracted local class centers decreased. As a result, the segmentation performance of LOGCAN++ is also degraded.
In addition, we specifically evaluate the contribution of the proposed affine transformer. Table \ref{tab:atb} compares the results achieved by our approach with and without the affine transform block (i.e., we remove all three affine transform factors). It can be observed that when the affine transform block is removed, the model performance decreases by 1.70 F1 value compared to the baseline and is lower than when only one affine transform factor is removed. The combined results show that ATB effectively improves the model segmentation performance.

\subsubsection{Analysis of the patch number}
We conduct experiments on the Vaihingen dataset to verify the effect of different number of local patches on the results (shown in Table \ref{tab:num}). Intuitively, the number of patches affects the size of the local window and thus the generation of local class centers. Since we split along the width and height of the feature map separately, we denote the patch number in the form of $N_h \times N_w$, e.g., $2\times 2$ means that we split the original feature map into four patches, while $1\times 1$ means that no split operation is performed, i.e., the global class center is used for context modeling in LCA module. It can be observed that as the number of patches increases, the performance of the model shows a general trend of increasing and then decreasing, and achieves an optimal value of 74.72\% at $4 \times4$. Therefore, we choose the number of patches to be $4 \times4$ in the final model.

\begin{table*}[t]
\centering
\caption{Ablation Study on the Impact of Head Numbers on the Vaihingen Dataset. The highest value is bolded.}
\setlength{\tabcolsep}{10pt}{
\begin{tabular}{c|c c c c c c|c}
\toprule
Head number &Impervious surface &Building & low vegetation &Tree & Car & Clutter &mIoU\\
\midrule
1 & \up{86.20}{0.33}&\down{91.57}{0.28}&\up{72.29}{0.85}&\up{80.61}{0.26}&\down{78.18}{1.96}&\down{32.33}{6.30}&\down{73.53}{1.19}\\
2 & \down{85.84}{0.03}&\down{91.69}{0.16}&\up{71.51}{0.07}&\up{80.52}{0.17}&\down{78.88}{1.26}&\down{33.83}{4.80}&\down{73.71}{1.01}\\
4 &\up{86.10}{0.23}&\down{91.80}{0.05}&\up{71.91}{0.47}&\up{80.68}{0.33}&\down{79.33}{0.81}&\down{35.08}{3.55}&\down{74.15}{0.57}\\
8 &85.87 &91.85 &71.44 &80.35 & 80.14 & 38.63 &74.72\\
16 &\up{86.20}{0.33} &\up{91.90}{0.05} &\up{72.01}{0.57} &\up{80.78}{0.43} &\down{79.43}{0.71} &\down{35.18}{3.45} &\down{74.25}{0.47}\\
    \bottomrule	
    \end{tabular}
}
\label{tab:head}
\end{table*}

\subsubsection{Analysis of the head number}
We investigate the effect caused by different numbers of heads on context modeling in the LCA module, where we set the number of heads as 1, 2, 4, 8 and 16 for the experiment. As shown in Table \ref{tab:head}, the number of heads increases, the performance of the model tends to increase and then decrease, and the phenomenon is similar to the multi-head self-attention in Transformer \cite{transformer}. This proves that by introducing the multi-head mechanism, LOGCAN++ is able to capture more complex pixel class relationships, thus improving the model performance.

\begin{table*}[t]
\centering
\caption{Ablation Study on the loss function coefficients on the Vaihingen Dataset. The highest value is bolded.}
\setlength{\tabcolsep}{10pt}{
\begin{tabular}{c|c c c c c c|c}
\toprule
Coefficient of  $\mathcal{L}_{ce}^{aux}$ &Impervious surface &Building & low vegetation &Tree & Car & Clutter &mIoU\\
\midrule
0.2 &\up{85.97}{0.10}&\down{91.60}{0.25}&\up{71.57}{0.13}&\down{80.26}{0.09}&\down{78.76}{1.38}&\down{35.14}{3.49}&\down{73.88}{0.84}\\
0.4 &\down{85.84}{0.03}&\down{91.50}{0.35}&\up{71.89}{0.45}&\up{80.67}{0.32}&\down{79.34}{0.80}&\down{36.74}{1.89}&\down{74.33}{0.39}\\
0.6 &\down{85.78}{0.09}&\down{91.60}{0.25}&\up{71.94}{0.50}&\up{80.43}{0.08}&\down{77.44}{2.70}&\up{39.57}{0.94}&\down{74.46}{0.26}\\
0.8 &85.87 &91.85 &71.44 &80.35 &80.14 &38.63 &74.72\\
1.0 &\down{85.74}{0.13}&\down{91.51}{0.34}&\up{71.50}{0.06}&\up{80.51}{0.16}&\down{78.58}{1.56}&\down{34.09}{4.54}&\down{73.65}{1.07}\\
    \bottomrule	
    \end{tabular}
}
\label{tab:loss}
\end{table*}

\subsubsection{Analysis of the loss function coefficients }
We finally explore the impact of different loss coefficients on model performance.
As shown in Table \ref{tab:loss}, when we set the weight of  auxiliary loss to 0.8, the proposed method achieves the the highest segmentation performance. Additionally, we observe that setting this value too low or too high slightly degrades the model's performance. 

\section{Conclusion}

\noindent In this paper, we begin with a detailed analysis of the characteristics of remote sensing images (i.e., complex background, scale and orientation variations, and high intra-class variance). Against this background, we propose the highly customized LOGCAN++ to cope with the segmentation challenges posed by the above characteristics. Specifically, LOGCAN++ is based on a local global class-aware strategy to reduce background noise interference and mitigate the intra-class variance problem. In particular, LOGCAN++ can tolerate scale and orientation variations in remote sensing images well due to the introduction of affine transformations in the process of extracting local class centers. Experimental results on three remote sensing semantic segmentation datasets show that LOGCAN++ significantly outperforms other state-of-the-art methods and possesses a better balance between accuracy and efficiency. 
In our future work, we will further explore it in conjunction with existing semantic segmentation macromodels such as SAM in order to fully utilize the potential of LOGCAN++ in the field of remote sensing image segmentation.


\bibliographystyle{IEEEtran}
\bibliography{IEEEabrv,myrefs}

\begin{thebibliography}{10}
\providecommand{\url}[1]{#1}
\csname url@samestyle\endcsname
\providecommand{\newblock}{\relax}
\providecommand{\bibinfo}[2]{#2}
\providecommand{\BIBentrySTDinterwordspacing}{\spaceskip=0pt\relax}
\providecommand{\BIBentryALTinterwordstretchfactor}{4}
\providecommand{\BIBentryALTinterwordspacing}{\spaceskip=\fontdimen2\font plus
\BIBentryALTinterwordstretchfactor\fontdimen3\font minus \fontdimen4\font\relax}
\providecommand{\BIBforeignlanguage}[2]{{%
\expandafter\ifx\csname l@#1\endcsname\relax
\typeout{** WARNING: IEEEtran.bst: No hyphenation pattern has been}%
\typeout{** loaded for the language `#1'. Using the pattern for}%
\typeout{** the default language instead.}%
\else
\language=\csname l@#1\endcsname
\fi
#2}}
\providecommand{\BIBdecl}{\relax}
\BIBdecl

\bibitem{environment}
Q.~Yuan, H.~Shen, T.~Li, Z.~Li, S.~Li, Y.~Jiang, H.~Xu, W.~Tan, Q.~Yang, J.~Wang \emph{et~al.}, ``Deep learning in environmental remote sensing: Achievements and challenges,'' \emph{Remote Sensing of Environment}, vol. 241, p. 111716, 2020.

\bibitem{water}
H.-F. Zhong, H.-M. Sun, D.-N. Han, Z.-H. Li, and R.-S. Jia, ``Lake water body extraction of optical remote sensing images based on semantic segmentation,'' \emph{Applied Intelligence}, pp. 1--16, 2022.

\bibitem{urban}
Q.~Zhang and K.~C. Seto, ``Mapping urbanization dynamics at regional and global scales using multi-temporal dmsp/ols nighttime light data,'' \emph{Remote Sensing of Environment}, vol. 115, no.~9, pp. 2320--2329, 2011.

\bibitem{urban2}
B.~Huang, B.~Zhao, and Y.~Song, ``Urban land-use mapping using a deep convolutional neural network with high spatial resolution multispectral remote sensing imagery,'' \emph{Remote Sensing of Environment}, vol. 214, pp. 73--86, 2018.

\bibitem{rsdataset}
F.~Bastani, P.~Wolters, R.~Gupta, J.~Ferdinando, and A.~Kembhavi, ``Satlaspretrain: A large-scale dataset for remote sensing image understanding,'' in \emph{Proceedings of the IEEE/CVF International Conference on Computer Vision (ICCV)}, October 2023, pp. 16\,772--16\,782.

\bibitem{9756442}
L.~Zhang and L.~Zhang, ``Artificial intelligence for remote sensing data analysis: A review of challenges and opportunities,'' \emph{IEEE Geoscience and Remote Sensing Magazine}, vol.~10, no.~2, pp. 270--294, 2022.

\bibitem{10542207}
M.~Lan, F.~Rong, H.~Jiao, Z.~Gao, and L.~Zhang, ``Language query-based transformer with multiscale cross-modal alignment for visual grounding on remote sensing images,'' \emph{IEEE Transactions on Geoscience and Remote Sensing}, vol.~62, pp. 1--13, 2024.

\bibitem{pspnet}
H.~Zhao, J.~Shi, X.~Qi, X.~Wang, and J.~Jia, ``Pyramid scene parsing network,'' in \emph{Proceedings of the IEEE conference on computer vision and pattern recognition}, 2017, pp. 2881--2890.

\bibitem{nonlocal}
X.~Wang, R.~Girshick, A.~Gupta, and K.~He, ``Non-local neural networks,'' in \emph{Proceedings of the IEEE conference on computer vision and pattern recognition}, 2018, pp. 7794--7803.

\bibitem{danet}
J.~Fu, J.~Liu, H.~Tian, Y.~Li, Y.~Bao, Z.~Fang, and H.~Lu, ``Dual attention network for scene segmentation,'' in \emph{Proceedings of the IEEE/CVF conference on computer vision and pattern recognition}, 2019, pp. 3146--3154.

\bibitem{ccnet}
Z.~Huang, X.~Wang, L.~Huang, C.~Huang, Y.~Wei, and W.~Liu, ``Ccnet: Criss-cross attention for semantic segmentation,'' in \emph{Proceedings of the IEEE/CVF international conference on computer vision}, 2019, pp. 603--612.

\bibitem{caa}
Y.~Huang, D.~Kang, W.~Jia, L.~Liu, and X.~He, ``Channelized axial attention--considering channel relation within spatial attention for semantic segmentation,'' in \emph{Proceedings of the AAAI Conference on Artificial Intelligence}, vol.~36, no.~1, 2022, pp. 1016--1025.

\bibitem{deeplabv3+}
L.-C. Chen, Y.~Zhu, G.~Papandreou, F.~Schroff, and H.~Adam, ``Encoder-decoder with atrous separable convolution for semantic image segmentation,'' in \emph{Proceedings of the European conference on computer vision (ECCV)}, 2018, pp. 801--818.

\bibitem{acfnet}
F.~Zhang, Y.~Chen, Z.~Li, Z.~Hong, J.~Liu, F.~Ma, J.~Han, and E.~Ding, ``Acfnet: Attentional class feature network for semantic segmentation,'' in \emph{Proceedings of the IEEE/CVF International Conference on Computer Vision}, 2019, pp. 6798--6807.

\bibitem{ocrnet}
Y.~Yuan, X.~Chen, and J.~Wang, ``Object-contextual representations for semantic segmentation,'' in \emph{Computer Vision--ECCV 2020: 16th European Conference, Glasgow, UK, August 23--28, 2020, Proceedings, Part VI 16}.\hskip 1em plus 0.5em minus 0.4em\relax Springer, 2020, pp. 173--190.

\bibitem{ccenet}
Q.~Wang, X.~Luo, J.~Feng, S.~Li, and J.~Yin, ``Ccenet: Cascade class-aware enhanced network for high-resolution aerial imagery semantic segmentation,'' \emph{IEEE Journal of Selected Topics in Applied Earth Observations and Remote Sensing}, vol.~15, pp. 6943--6956, 2022.

\bibitem{hmanet}
R.~Niu, X.~Sun, Y.~Tian, W.~Diao, K.~Chen, and K.~Fu, ``Hybrid multiple attention network for semantic segmentation in aerial images,'' \emph{IEEE Transactions on Geoscience and Remote Sensing}, vol.~60, pp. 1--18, 2021.

\bibitem{segmenter}
R.~Strudel, R.~Garcia, I.~Laptev, and C.~Schmid, ``Segmenter: Transformer for semantic segmentation,'' in \emph{Proceedings of the IEEE/CVF international conference on computer vision}, 2021, pp. 7262--7272.

\bibitem{segformer}
E.~Xie, W.~Wang, Z.~Yu, A.~Anandkumar, J.~M. Alvarez, and P.~Luo, ``Segformer: Simple and efficient design for semantic segmentation with transformers,'' \emph{Advances in Neural Information Processing Systems}, vol.~34, pp. 12\,077--12\,090, 2021.

\bibitem{unetformer}
L.~Wang, R.~Li, C.~Zhang, S.~Fang, C.~Duan, X.~Meng, and P.~M. Atkinson, ``Unetformer: A unet-like transformer for efficient semantic segmentation of remote sensing urban scene imagery,'' \emph{ISPRS Journal of Photogrammetry and Remote Sensing}, vol. 190, pp. 196--214, 2022.

\bibitem{mask2former}
B.~Cheng, I.~Misra, A.~G. Schwing, A.~Kirillov, and R.~Girdhar, ``Masked-attention mask transformer for universal image segmentation,'' in \emph{Proceedings of the IEEE/CVF conference on computer vision and pattern recognition}, 2022, pp. 1290--1299.

\bibitem{biformer}
L.~Zhu, X.~Wang, Z.~Ke, W.~Zhang, and R.~W. Lau, ``Biformer: Vision transformer with bi-level routing attention,'' in \emph{Proceedings of the IEEE/CVF Conference on Computer Vision and Pattern Recognition}, 2023, pp. 10\,323--10\,333.

\bibitem{logcan}
X.~Ma, M.~Ma, C.~Hu, Z.~Song, Z.~Zhao, T.~Feng, and W.~Zhang, ``Log-can: local-global class-aware network for semantic segmentation of remote sensing images,'' in \emph{ICASSP 2023-2023 IEEE International Conference on Acoustics, Speech and Signal Processing (ICASSP)}.\hskip 1em plus 0.5em minus 0.4em\relax IEEE, 2023, pp. 1--5.

\bibitem{fcn}
J.~Long, E.~Shelhamer, and T.~Darrell, ``Fully convolutional networks for semantic segmentation,'' in \emph{Proceedings of the IEEE conference on computer vision and pattern recognition}, 2015, pp. 3431--3440.

\bibitem{semantic}
L.-C. Chen, G.~Papandreou, I.~Kokkinos, K.~Murphy, and A.~L. Yuille, ``Semantic image segmentation with deep convolutional nets and fully connected crfs,'' \emph{arXiv preprint arXiv:1412.7062}, 2014.

\bibitem{Deeplab}
{Chen, Liang-Chieh and Papandreou, George and Kokkinos, Iasonas and Murphy, Kevin and Yuille, Alan L}, ``Deeplab: Semantic image segmentation with deep convolutional nets, atrous convolution, and fully connected crfs,'' \emph{IEEE transactions on pattern analysis and machine intelligence}, vol.~40, no.~4, pp. 834--848, 2017.

\bibitem{deeplabv2}
L.-C. Chen, G.~Papandreou, F.~Schroff, and H.~Adam, ``Rethinking atrous convolution for semantic image segmentation,'' \emph{arXiv preprint arXiv:1706.05587}, 2017.

\bibitem{denseaspp}
M.~Yang, K.~Yu, C.~Zhang, Z.~Li, and K.~Yang, ``Denseaspp for semantic segmentation in street scenes,'' in \emph{Proceedings of the IEEE conference on computer vision and pattern recognition}, 2018, pp. 3684--3692.

\bibitem{dmnet}
J.~He, Z.~Deng, and Y.~Qiao, ``Dynamic multi-scale filters for semantic segmentation,'' in \emph{Proceedings of the IEEE/CVF International Conference on Computer Vision}, 2019, pp. 3562--3572.

\bibitem{flanet}
Q.~Song, J.~Li, C.~Li, H.~Guo, and R.~Huang, ``Fully attentional network for semantic segmentation,'' in \emph{Proceedings of the AAAI Conference on Artificial Intelligence}, vol.~36, no.~2, 2022, pp. 2280--2288.

\bibitem{dmsanet}
A.~Sagar, ``Dmsanet: Dual multi scale attention network,'' in \emph{International Conference on Image Analysis and Processing}.\hskip 1em plus 0.5em minus 0.4em\relax Springer, 2022, pp. 633--645.

\bibitem{ranet}
D.~Shen, Y.~Ji, P.~Li, Y.~Wang, and D.~Lin, ``Ranet: Region attention network for semantic segmentation,'' \emph{Advances in Neural Information Processing Systems}, vol.~33, pp. 13\,927--13\,938, 2020.

\bibitem{cpnet}
C.~Yu, J.~Wang, C.~Gao, G.~Yu, C.~Shen, and N.~Sang, ``Context prior for scene segmentation,'' in \emph{Proceedings of the IEEE/CVF conference on computer vision and pattern recognition}, 2020, pp. 12\,416--12\,425.

\bibitem{docnet}
X.~Ma, R.~Che, X.~Wang, M.~Ma, S.~Wu, T.~Feng, and W.~Zhang, ``Docnet: Dual-domain optimized class-aware network for remote sensing image segmentation,'' \emph{IEEE Geoscience and Remote Sensing Letters}, 2024.

\bibitem{isnet}
Z.~Jin, B.~Liu, Q.~Chu, and N.~Yu, ``Isnet: Integrate image-level and semantic-level context for semantic segmentation,'' in \emph{Proceedings of the IEEE/CVF International Conference on Computer Vision}, 2021, pp. 7189--7198.

\bibitem{transformer}
A.~Vaswani, N.~Shazeer, N.~Parmar, J.~Uszkoreit, L.~Jones, A.~N. Gomez, {\L}.~Kaiser, and I.~Polosukhin, ``Attention is all you need,'' \emph{Advances in neural information processing systems}, vol.~30, 2017.

\bibitem{beit}
H.~Bao, L.~Dong, S.~Piao, and F.~Wei, ``Beit: Bert pre-training of image transformers,'' \emph{arXiv preprint arXiv:2106.08254}, 2021.

\bibitem{detr}
N.~Carion, F.~Massa, G.~Synnaeve, N.~Usunier, A.~Kirillov, and S.~Zagoruyko, ``End-to-end object detection with transformers,'' in \emph{European conference on computer vision}.\hskip 1em plus 0.5em minus 0.4em\relax Springer, 2020, pp. 213--229.

\bibitem{setr}
S.~Zheng, J.~Lu, H.~Zhao, X.~Zhu, Z.~Luo, Y.~Wang, Y.~Fu, J.~Feng, T.~Xiang, P.~H. Torr, and L.~Zhang, ``Rethinking semantic segmentation from a sequence-to-sequence perspective with transformers,'' in \emph{Proceedings of the IEEE/CVF Conference on Computer Vision and Pattern Recognition (CVPR)}, June 2021, pp. 6881--6890.

\bibitem{botnet}
A.~Srinivas, T.-Y. Lin, N.~Parmar, J.~Shlens, P.~Abbeel, and A.~Vaswani, ``Bottleneck transformers for visual recognition,'' in \emph{Proceedings of the IEEE/CVF conference on computer vision and pattern recognition}, 2021, pp. 16\,519--16\,529.

\bibitem{deit}
H.~Touvron, M.~Cord, M.~Douze, F.~Massa, A.~Sablayrolles, and H.~J{\'e}gou, ``Training data-efficient image transformers \& distillation through attention,'' in \emph{International conference on machine learning}.\hskip 1em plus 0.5em minus 0.4em\relax PMLR, 2021, pp. 10\,347--10\,357.

\bibitem{vit}
A.~Dosovitskiy, L.~Beyer, A.~Kolesnikov, D.~Weissenborn, X.~Zhai, T.~Unterthiner, M.~Dehghani, M.~Minderer, G.~Heigold, S.~Gelly \emph{et~al.}, ``An image is worth 16x16 words: Transformers for image recognition at scale,'' \emph{arXiv preprint arXiv:2010.11929}, 2020.

\bibitem{swintransformer}
Z.~Liu, Y.~Lin, Y.~Cao, H.~Hu, Y.~Wei, Z.~Zhang, S.~Lin, and B.~Guo, ``Swin transformer: Hierarchical vision transformer using shifted windows,'' in \emph{Proceedings of the IEEE/CVF international conference on computer vision}, 2021, pp. 10\,012--10\,022.

\bibitem{maskformer}
B.~Cheng, A.~Schwing, and A.~Kirillov, ``Per-pixel classification is not all you need for semantic segmentation,'' \emph{Advances in Neural Information Processing Systems}, vol.~34, pp. 17\,864--17\,875, 2021.

\bibitem{mpformer}
H.~Zhang, F.~Li, H.~Xu, S.~Huang, S.~Liu, L.~M. Ni, and L.~Zhang, ``Mp-former: Mask-piloted transformer for image segmentation,'' in \emph{Proceedings of the IEEE/CVF Conference on Computer Vision and Pattern Recognition (CVPR)}, June 2023, pp. 18\,074--18\,083.

\bibitem{faseg}
H.~He, J.~Cai, Z.~Pan, J.~Liu, J.~Zhang, D.~Tao, and B.~Zhuang, ``Dynamic focus-aware positional queries for semantic segmentation,'' in \emph{Proceedings of the IEEE/CVF Conference on Computer Vision and Pattern Recognition (CVPR)}, June 2023, pp. 11\,299--11\,308.

\bibitem{pem}
N.~Cavagnero, G.~Rosi, C.~Ruttano, F.~Pistilli, M.~Ciccone, G.~Averta, and F.~Cermelli, ``Pem: Prototype-based efficient maskformer for image segmentation,'' \emph{arXiv preprint arXiv:2402.19422}, 2024.

\bibitem{ddp}
Y.~Ji, Z.~Chen, E.~Xie, L.~Hong, X.~Liu, Z.~Liu, T.~Lu, Z.~Li, and P.~Luo, ``Ddp: Diffusion model for dense visual prediction,'' in \emph{Proceedings of the IEEE/CVF International Conference on Computer Vision}, 2023, pp. 21\,741--21\,752.

\bibitem{land1}
C.~Zhang, I.~Sargent, X.~Pan, H.~Li, A.~Gardiner, J.~Hare, and P.~M. Atkinson, ``An object-based convolutional neural network (ocnn) for urban land use classification,'' \emph{Remote sensing of environment}, vol. 216, pp. 57--70, 2018.

\bibitem{jdp}
{Ce Zhang and Isabel Sargent and Xin Pan and Huapeng Li and Andy Gardiner and Jonathon Hare and Peter M. Atkinson}, ``Joint deep learning for land cover and land use classification,'' \emph{Remote Sensing of Environment}, vol. 221, pp. 173--187, 2019.

\bibitem{building1}
H.~Jung, H.-S. Choi, and M.~Kang, ``Boundary enhancement semantic segmentation for building extraction from remote sensed image,'' \emph{IEEE Transactions on Geoscience and Remote Sensing}, vol.~60, pp. 1--12, 2021.

\bibitem{lbe}
J.~Yuan, ``Learning building extraction in aerial scenes with convolutional networks,'' \emph{IEEE transactions on pattern analysis and machine intelligence}, vol.~40, no.~11, pp. 2793--2798, 2017.

\bibitem{building2}
K.~Lee, J.~H. Kim, H.~Lee, J.~Park, J.~P. Choi, and J.~Y. Hwang, ``Boundary-oriented binary building segmentation model with two scheme learning for aerial images,'' \emph{IEEE Transactions on Geoscience and Remote Sensing}, vol.~60, pp. 1--17, 2021.

\bibitem{bdtnet}
L.~Luo, J.-X. Wang, S.-B. Chen, J.~Tang, and B.~Luo, ``Bdtnet: Road extraction by bi-direction transformer from remote sensing images,'' \emph{IEEE Geoscience and Remote Sensing Letters}, vol.~19, pp. 1--5, 2022.

\bibitem{bmda}
S.~Dong and Z.~Chen, ``Block multi-dimensional attention for road segmentation in remote sensing imagery,'' \emph{IEEE Geoscience and Remote Sensing Letters}, vol.~19, pp. 1--5, 2021.

\bibitem{roade2}
Y.~Wang, Y.~Peng, W.~Li, G.~C. Alexandropoulos, J.~Yu, D.~Ge, and W.~Xiang, ``Ddu-net: dual-decoder-u-net for road extraction using high-resolution remote sensing images,'' \emph{IEEE Transactions on Geoscience and Remote Sensing}, vol.~60, pp. 1--12, 2022.

\bibitem{roadtracer}
F.~Bastani, S.~He, S.~Abbar, M.~Alizadeh, H.~Balakrishnan, S.~Chawla, S.~Madden, and D.~DeWitt, ``Roadtracer: Automatic extraction of road networks from aerial images,'' in \emph{Proceedings of the IEEE conference on computer vision and pattern recognition}, 2018, pp. 4720--4728.

\bibitem{vis}
L.~Mou and X.~X. Zhu, ``Vehicle instance segmentation from aerial image and video using a multitask learning residual fully convolutional network,'' \emph{IEEE Transactions on Geoscience and Remote Sensing}, vol.~56, no.~11, pp. 6699--6711, 2018.

\bibitem{lanet}
L.~Ding, H.~Tang, and L.~Bruzzone, ``Lanet: Local attention embedding to improve the semantic segmentation of remote sensing images,'' \emph{IEEE Transactions on Geoscience and Remote Sensing}, vol.~59, no.~1, pp. 426--435, 2020.

\bibitem{manet}
R.~Li, S.~Zheng, C.~Zhang, C.~Duan, J.~Su, L.~Wang, and P.~M. Atkinson, ``Multiattention network for semantic segmentation of fine-resolution remote sensing images,'' \emph{IEEE Transactions on Geoscience and Remote Sensing}, vol.~60, pp. 1--13, 2021.

\bibitem{pointflow}
X.~Li, H.~He, X.~Li, D.~Li, G.~Cheng, J.~Shi, L.~Weng, Y.~Tong, and Z.~Lin, ``Pointflow: Flowing semantics through points for aerial image segmentation,'' in \emph{Proceedings of the IEEE/CVF Conference on Computer Vision and Pattern Recognition}, 2021, pp. 4217--4226.

\bibitem{sco}
F.~Yang and C.~Ma, ``Sparse and complete latent organization for geospatial semantic segmentation,'' in \emph{Proceedings of the IEEE/CVF Conference on Computer Vision and Pattern Recognition}, 2022, pp. 1809--1818.

\bibitem{glots}
Y.~Liu, Y.~Zhang, Y.~Wang, and S.~Mei, ``Rethinking transformers for semantic segmentation of remote sensing images,'' \emph{IEEE Transactions on Geoscience and Remote Sensing}, 2023.

\bibitem{farseg++}
Z.~Zheng, Y.~Zhong, J.~Wang, A.~Ma, and L.~Zhang, ``Farseg++: Foreground-aware relation network for geospatial object segmentation in high spatial resolution remote sensing imagery,'' \emph{IEEE Transactions on Pattern Analysis and Machine Intelligence}, 2023.

\bibitem{farseg}
Z.~Zheng, Y.~Zhong, J.~Wang, and A.~Ma, ``Foreground-aware relation network for geospatial object segmentation in high spatial resolution remote sensing imagery,'' in \emph{Proceedings of the IEEE/CVF conference on computer vision and pattern recognition}, 2020, pp. 4096--4105.

\bibitem{mdanet}
R.~Zuo, G.~Zhang, R.~Zhang, and X.~Jia, ``A deformable attention network for high-resolution remote sensing images semantic segmentation,'' \emph{IEEE Transactions on Geoscience and Remote Sensing}, vol.~60, pp. 1--14, 2022.

\bibitem{fpn}
A.~Kirillov, R.~Girshick, K.~He, and P.~Doll{\'a}r, ``Panoptic feature pyramid networks,'' in \emph{Proceedings of the IEEE/CVF conference on computer vision and pattern recognition}, 2019, pp. 6399--6408.

\bibitem{dcswin}
L.~Wang, R.~Li, C.~Duan, C.~Zhang, X.~Meng, and S.~Fang, ``A novel transformer based semantic segmentation scheme for fine-resolution remote sensing images,'' \emph{IEEE Geoscience and Remote Sensing Letters}, vol.~19, pp. 1--5, 2022.

\bibitem{EfficientViT}
H.~Cai, J.~Li, M.~Hu, C.~Gan, and S.~Han, ``Efficientvit: Multi-scale linear attention for high-resolution dense prediction,'' \emph{arXiv preprint arXiv:2205.14756}, 2022.

\bibitem{10026298}
R.~Xu, C.~Wang, J.~Zhang, S.~Xu, W.~Meng, and X.~Zhang, ``Rssformer: Foreground saliency enhancement for remote sensing land-cover segmentation,'' \emph{IEEE Transactions on Image Processing}, vol.~32, pp. 1052--1064, 2023.

\bibitem{tsne}
L.~Van~der Maaten and G.~Hinton, ``Visualizing data using t-sne.'' \emph{Journal of machine learning research}, vol.~9, no.~11, 2008.

\bibitem{www.isprs.org}
F.~Rottensteiner, G.~Sohn, J.~Jung, M.~Gerke, C.~Baillard, S.~Bnitez, and U.~Breitkopf, ``International society for photogrammetry and remote sensing, 2d semantic labeling contest,'' Accessed: Oct. 29, 2020., available: \url{https://www.isprs.org/education/benchmarks}.

\bibitem{loveda}
J.~Wang, Z.~Zheng, A.~Ma, X.~Lu, and Y.~Zhong, ``Loveda: A remote sensing land-cover dataset for domain adaptive semantic segmentation,'' \emph{arXiv preprint arXiv:2110.08733}, 2021.

\bibitem{resnet}
K.~He, X.~Zhang, S.~Ren, and J.~Sun, ``Deep residual learning for image recognition,'' in \emph{Proceedings of the IEEE conference on computer vision and pattern recognition}, 2016, pp. 770--778.

\bibitem{gradcam}
R.~R. Selvaraju, M.~Cogswell, A.~Das, R.~Vedantam, D.~Parikh, and D.~Batra, ``Grad-cam: Visual explanations from deep networks via gradient-based localization,'' in \emph{Proceedings of the IEEE international conference on computer vision}, 2017, pp. 618--626.

\end{thebibliography}

\vfill

\end{document}